\documentclass[runningheads]{llncs}
\usepackage{graphicx}

\usepackage{times}
\usepackage{epsfig}
\usepackage{amsmath}
\usepackage{amssymb}

\usepackage{pdfpages} 

\usepackage{mathtools}
\usepackage{fancyhdr}
\usepackage{url}
\usepackage{enumitem}
\usepackage{subcaption}
\usepackage{multirow}
\usepackage{epstopdf}

\begin{document}

\title{ReGenMorph: Visibly Realistic GAN Generated Face Morphing Attacks by Attack Re-generation}
%
\titlerunning{ReGenMorph: Visibly Realistic GAN Generated Face Morphing Attacks \thanks{Supported by organization x.}}
%
\author{Naser Damer \inst{1,2} \and Kiran Raja \inst{3} \and Marius S\"{u}ßmilch \inst{1}  \and Sushma Venkatesh \inst{3}  \and \\ Fadi Boutros \inst{1,2} \and Meiling Fang \inst{1,2} \and Florian Kirchbuchner \inst{1} \and \\ Raghavendra Ramachandra \inst{3} \and Arjan Kuijper\inst{1,2}
}

\authorrunning{N. Damer et al.}

\institute{Fraunhofer Institute for Computer Graphics Research IGD, Germany \and
Department of Computer Science, TU Darmstadt, Germany
\and
Norwegian University of Science and Technology, Norway \\
\email{naser.damer@igd.fraunhofer.de}\\}

\maketitle              

\begin{abstract}

Face morphing attacks aim at creating face images that are verifiable to be the face of multiple identities, which can lead to building faulty identity links in operations like border checks. While creating a morphed face detector (MFD), training on all possible attack types is essential to achieve good detection performance. Therefore, investigating new methods of creating morphing attacks drives the generalizability of MADs.
Creating morphing attacks was performed on the image level, by landmark interpolation, or on the latent-space level, by manipulating latent vectors in a generative adversarial network. The earlier results in varying blending artifacts and the latter results in synthetic-like striping artifacts. 
This work presents the novel morphing pipeline, ReGenMorph, to eliminate the LMA blending artifacts by using a GAN-based generation, as well as, eliminate the manipulation in the latent space, resulting in visibly realistic morphed images compared to previous works. 
The generated ReGenMorph appearance is compared to recent morphing approaches and evaluated for face recognition vulnerability and attack detectability, whether as known or unknown attacks.

\end{abstract}
%
%
%

\section{Introduction}
\label{sec:int}

The deep-learning driven performance advances in face recognition \cite{DBLP:conf/cvpr/DengGXZ19}, along with the relatively high social acceptance \cite{Jain:1998:BPI:552539}, have brought automatic face recognition to be a key technology in security sensitive applications of identity management (e.g. travel documents) \cite{FaceMarket2017}. However, face recognition systems are vulnerable to many attacks, one of these is face morphing attacks.
Ferrara et al. \cite{DBLP:conf/icb/FerraraFM14} discussed the face morphing attack by showing that one face reference attack image can be, automatically and by human experts, successfully matched to more than one person. If such morphing attacks are used in travel or identity documents, it would allow multiple subjects to verify their identity to the one associated with the document. This faulty subject link to the document identity can lead to a wide range of illegal activities, including financial transactions, illegal immigration, human trafficking, and circumventing criminal identity lists. 

The first proposed morphing attacks were morphed on the image level. These are commonly created by interpolating facial landmarks in the morphed images and blending the texture information \cite{DBLP:conf/icb/RaghavendraRVB17,DBLP:journals/tifs/FerraraFM18}, i.e. landmark-based attacks (LMA). However, this image-level interpolation commonly causes blending image artifacts. 
Knowing (during training) possible novel approaches of creating morphing attacks is essential to create generalizable morphing attack detectors (MAD) \cite{DBLP:conf/icb/DamerSZWTKK19}. 
Motivated by that, several researchers proposed to take advantage of the ever-increasing capabilities of the generative adversarial networks (GAN) to generate face morphing attacks \cite{DBLP:conf/btas/DamerS0K18,DBLP:conf/btas/DamerBSKK19,DBLP:conf/iwbf/VenkateshZRRDB20,DBLP:journals/corr/abs-2009-01729MIPGAN}. These works performed the identity interpolation on the latent vector level, rather than the image-level in the LMA approaches.
However, manipulating the latent vector resulted in generated images with slight synthetic-like striping artifacts, even in the most recent methods \cite{DBLP:journals/corr/abs-2009-01729MIPGAN}.

In this work, we propose a novel face morphing concept, the ReGenMorphand, to avoid both, the blending artifacts in LMA and the synthetic striping artifacts in GAN-based morphs.
The proposed ReGenMorph approach achieves that by using a GAN-based generation, as well as, eliminate the manipulation in the latent space. This is done by performing the identity interpolation on the image level (just as LMA) but passing the image into a fine-tuned StylGAN encoder and generator, without latent space manipulation. This results in a generated morphed image of high quality and low visible artifacts, that also holds the identity information of the blended LMA image.
This paper presents our ReGenMorph approach in detail. We present a face recognition vulnerability study in comparison to LMA and the latest GAN-based attacks. We also study the detectability of the ReGenMorph attacks as known and unknown attacks using two of the top-performing MAD. Samples of the ReGenMorph images are presented, along with LMA and recent GAN-based attacks, for visual comparison.
In details, our experiments compare our ReGenMorphs to 3 morphing approached from \cite{DBLP:conf/icb/RaghavendraRVB17} (IJCB2017), \cite{DBLP:conf/iwbf/VenkateshZRRDB20} (IWBF2020), and \cite{DBLP:journals/corr/abs-2009-01729MIPGAN} (T-BIOM2021). We studied the vulnerability of all the considered attack types on both, top academic solution (ArcFace \cite{DBLP:conf/cvpr/DengGXZ19}) and one of the top COTS available \cite{cognitecfrssdk}. We evaluated the detectability of the presented morphs in known and unknown settings using two detection methods \cite{DBLP:conf/isba/RamachandraVRB19} (ISBA2019) and \cite{DBLP:conf/fusion/VenkateshRRB20} (FUSION2020) that achieved top performances in the morphing detection NIST challenge \cite{nistMorph}. 


\section{Related Works}
\label{sec:rw}

Morphing face images was initially performed by detecting facial landmarks in the source image to be morphed. These landmarks are later interpolated and the texture is blended to produce the morphed face image (LMA morphs). Slightly different versions of this process were used in different work, such as the work of Ferrera et al. \cite{DBLP:journals/tifs/FerraraFM18} and Ramachandra et al. \cite{DBLP:conf/icb/RaghavendraRVB17}. A comparison \cite{DBLP:journals/iet-bmt/ScherhagKRB20} of these methods have shown that the approach used in \cite{DBLP:conf/icb/RaghavendraRVB17} and \cite{DBLP:conf/dagm/DamerBWBTBK18} achieved the strongest face morphing attacks, i.e. highest identity preservation of the morphed identities. A more sophisticated variation of this process applied the interpolation on partial parts of the face \cite{DBLP:journals/tbbis/QinPVRLB21}, resulting in attacks that are harder to detect when they are not known to the morphing attack detector (MAD). 
The listed LMA morphs have various degrees of image artifacts introduced by the fact that the identity interpolation is performed on the image-level \cite{DBLP:journals/corr/abs-2009-01729MIPGAN}. 

Taking advantage of the advanced GAN architectures and their ability to produce synthetic images, and to avoid the image-level interpolation, Damer et al. proposed the MorGAN GAN-based morphing approach \cite{DBLP:conf/btas/DamerS0K18}. The MorGAN transferred the original images to be morphed into the latent space of the GAN and performed a latent-level interpolation. The interpolated latent vector is then used by the GAN generator to generate the morphed face image. These MorGAN images preserved the identities moderately and had low resolution. The MorGAN attacks have proven to be hard to detect if they were unknown to the MAD \cite{DBLP:conf/fusion/DamerZWSKK19,DBLP:conf/btas/DamerGZKK19}. A follow-up work added a post-generation cascaded enhancement step on the MorGAN network to increase the image quality, however, with the same identity preservation qualities \cite{DBLP:conf/btas/DamerBSKK19}.
Based on the idea of latent vector interpolation introduced in \cite{DBLP:conf/btas/DamerS0K18}, Venkatesh et al. created much more realistic and higher quality morphed images with better identity preservation qualities \cite{DBLP:conf/iwbf/VenkateshZRRDB20}. This advancement in \cite{DBLP:conf/iwbf/VenkateshZRRDB20} was mainly due to the use of an advanced GAN architecture, namely the StyleGAN by Karras et al. \cite{DBLP:conf/cvpr/KarrasLA19}. Also based on the StyleGAN architecture \cite{DBLP:conf/cvpr/KarrasLA19}, the MIPGAN-II generative morphing approach was introduced to generate images with higher identity preservation \cite{DBLP:journals/corr/abs-2009-01729MIPGAN}. This was achieved in \cite{DBLP:journals/corr/abs-2009-01729MIPGAN} by introducing a loss to optimize the identity preservation in the latent vector.

Although the existing LMA morphs have strong identity preservation capabilities, the fact that they build their identity blend on the image level makes them prone to image artifacts. The GAN-based morphs do generally preserve the identity to a lower degree \cite{DBLP:journals/corr/abs-2009-01729MIPGAN}, however, this is less relevant in such a scenario where the worst-case attack scenario needs to be considered. Despite the enhanced quality of the GAN-based morphed images, the manipulation in the latent space still produce synthetic-like generation artifacts \cite{DBLP:conf/btas/DamerS0K18,DBLP:conf/iwbf/VenkateshZRRDB20,DBLP:journals/corr/abs-2009-01729MIPGAN}. This work aims to eliminate the LMA blending artifacts by using a GAN-based generation, as well as, eliminate the manipulation in the latent space, resulting in visibly realistic morphed images compared to previous works.

\section{Methodology}
\label{sec:meh}

\begin{figure*}[ht]
         \centering
         \includegraphics[width=0.95\textwidth]{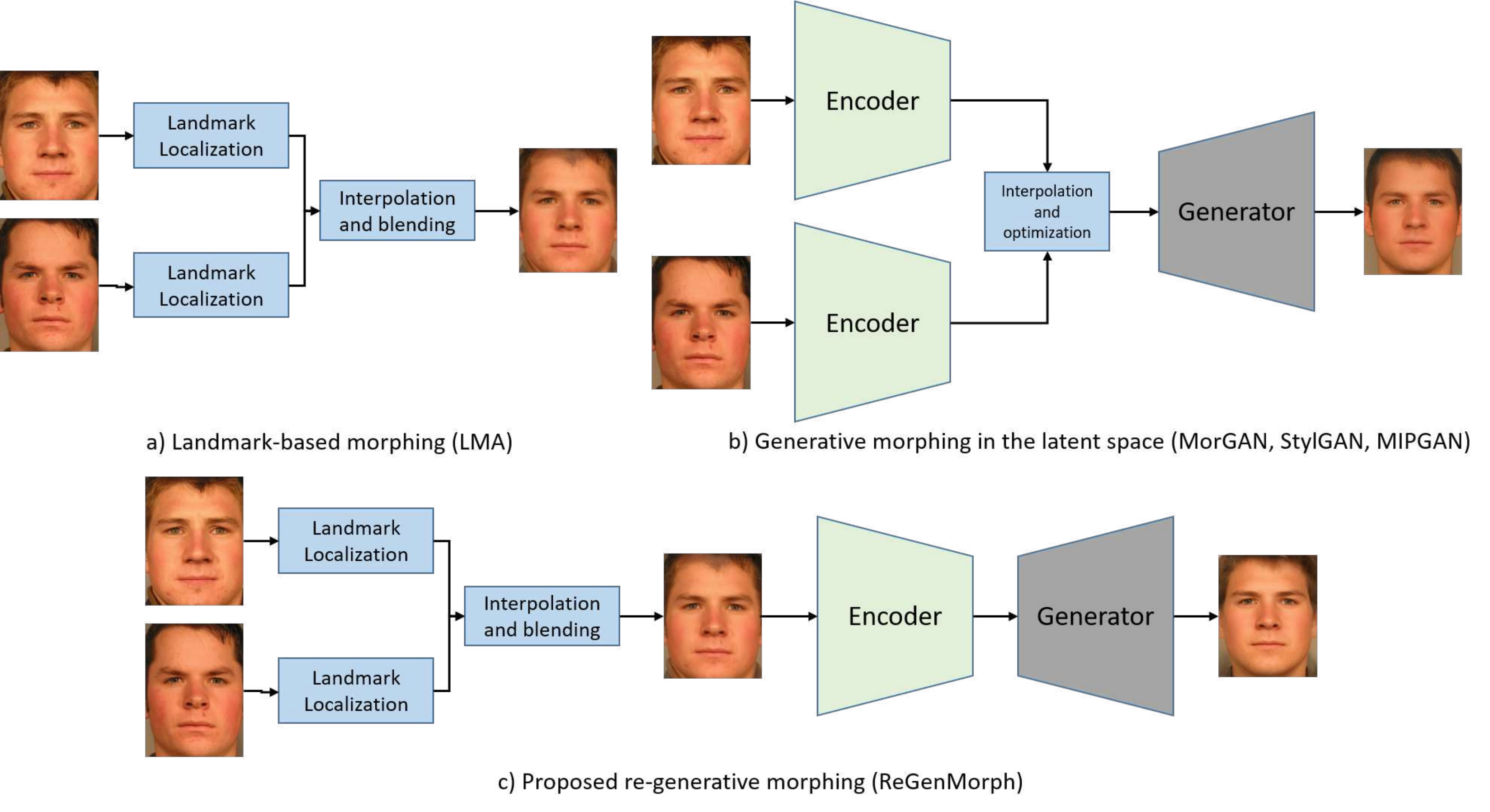}
        \caption{The face morphing pipelines using (a) facial landmarks - LMA \cite{DBLP:conf/icb/RaghavendraRVB17,DBLP:journals/tifs/FerraraFM18}, (b) conventional GAN-based morphing \cite{DBLP:conf/btas/DamerBSKK19,DBLP:conf/btas/DamerS0K18,DBLP:journals/corr/abs-2009-01729MIPGAN,DBLP:conf/iwbf/VenkateshZRRDB20}, and (c) the proposed ReGenMorph pipeline.}
        \label{fig:pipelines}
\end{figure*}

This section introduces our novel face image morphing pipeline and the technical details of our adaption to this pipeline.

\subsection{The ReGenMorph face morphing pipeline}
\begin{figure*}[t]
     \centering
     \begin{subfigure}[b]{0.45\textwidth}
         \centering
         \includegraphics[height=0.9\textwidth]{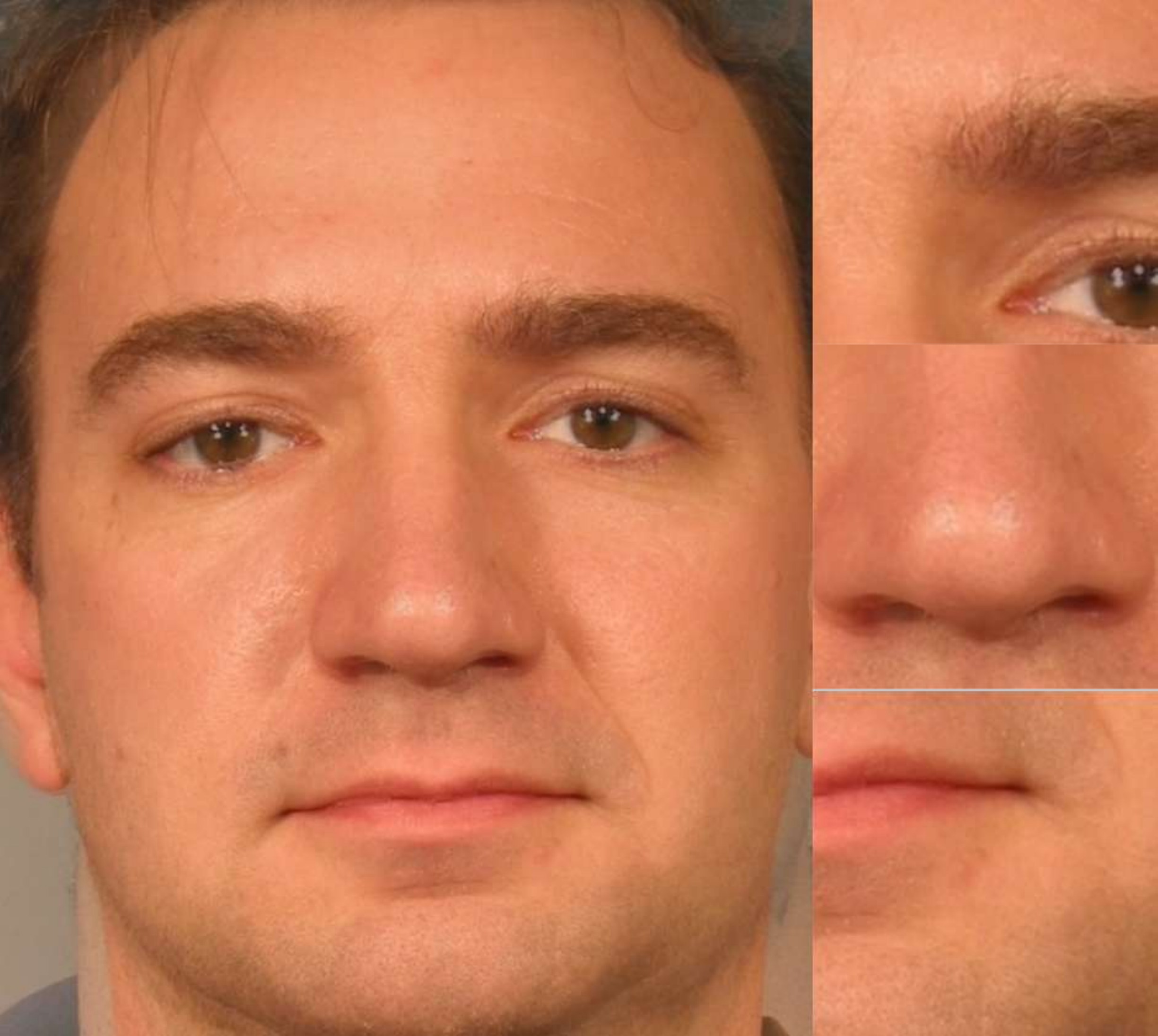}
         \caption{LMA \cite{DBLP:conf/icb/RaghavendraRVB17}}
         \label{fig:samp:BL}
     \end{subfigure}
     \hspace{8mm}
     \begin{subfigure}[b]{0.45\textwidth}
         \centering
         \includegraphics[height=0.9\textwidth]{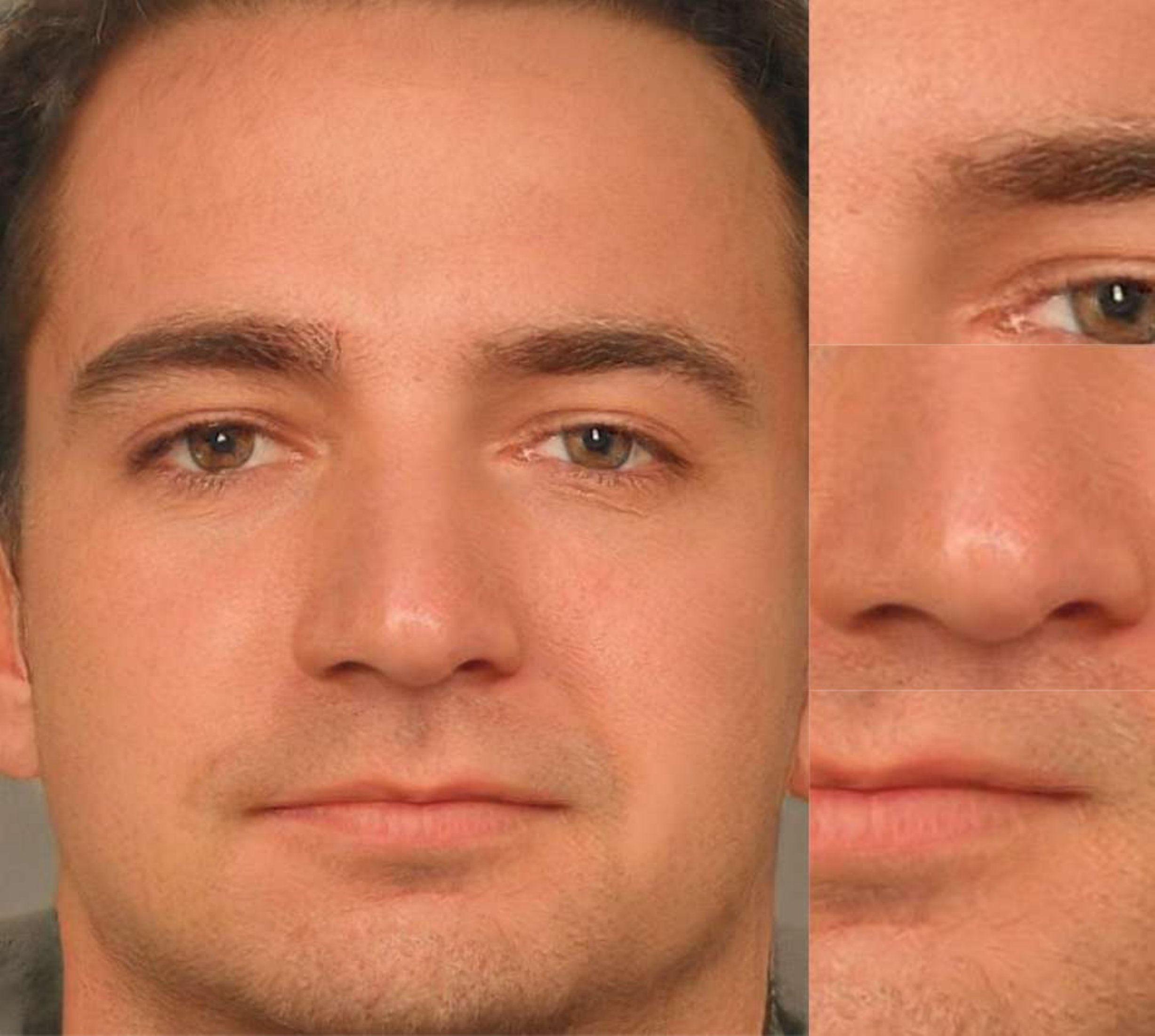}
         \caption{StyleGAN \cite{DBLP:conf/iwbf/VenkateshZRRDB20}}
         \label{fig:samp:BL}
     \end{subfigure}

     \begin{subfigure}[b]{0.45\textwidth}
         \centering
         \includegraphics[height=0.9\textwidth]{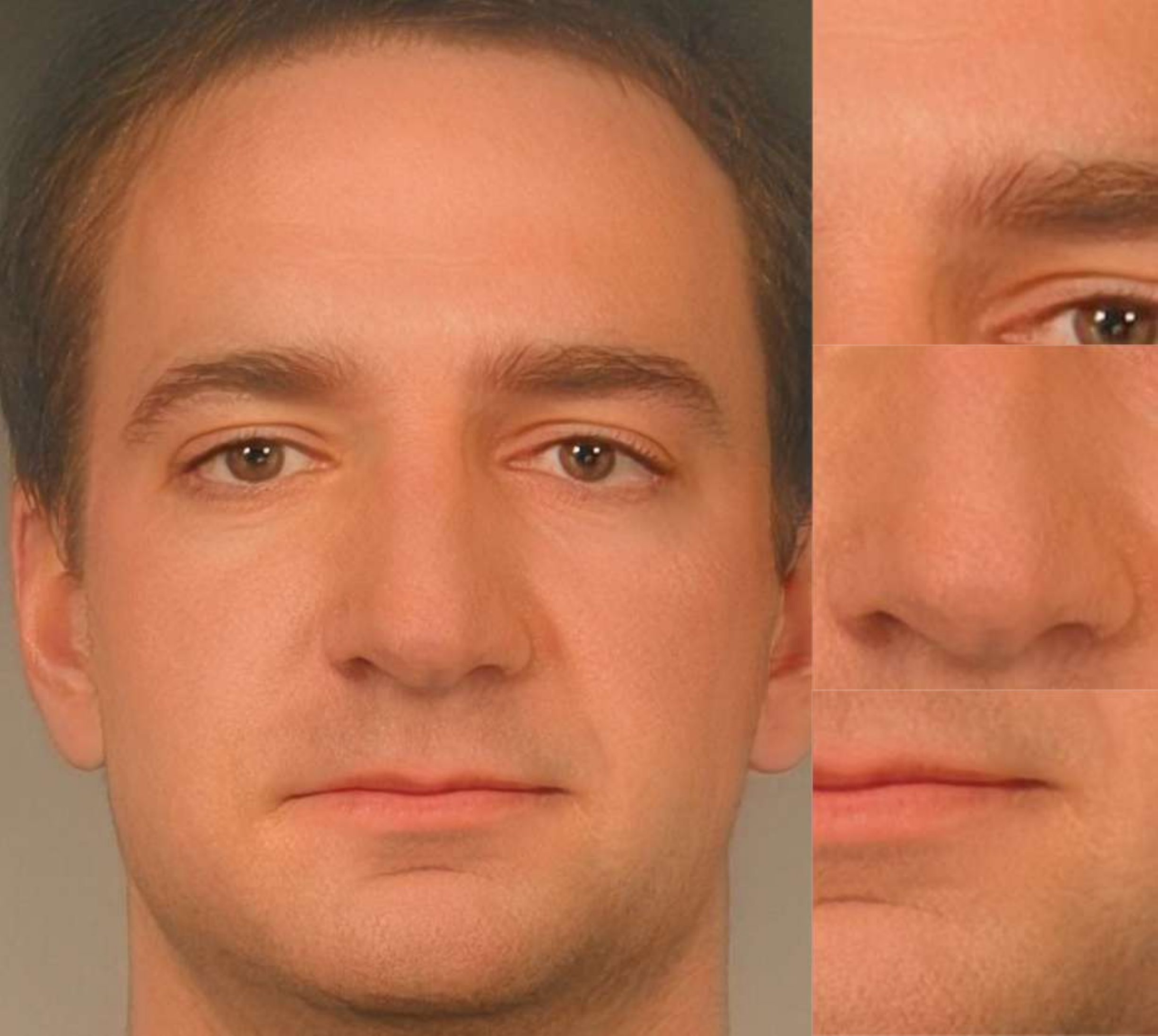}
         \caption{MIPGAN-II \cite{DBLP:journals/corr/abs-2009-01729MIPGAN}}
         \label{fig:samp:BL}
     \end{subfigure}
     \hspace{8mm}
     \begin{subfigure}[b]{0.45\textwidth}
         \centering
         \includegraphics[height=0.9\textwidth]{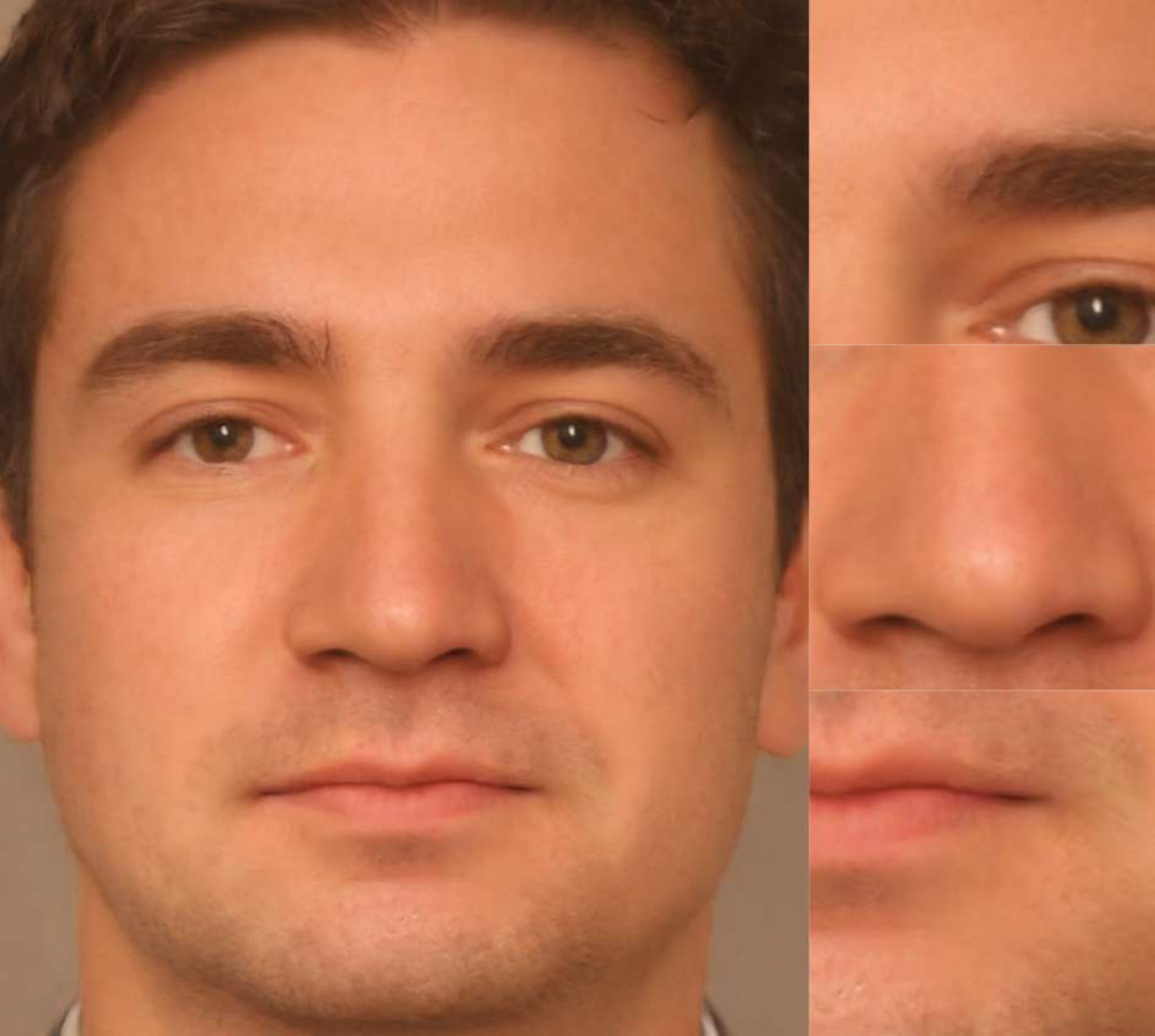}
         \caption{ReGenMorph (ours)}
         \label{fig:samp:BL}
     \end{subfigure}
        \caption{A large scale examples of a morphed image using LMA \cite{DBLP:conf/icb/RaghavendraRVB17}, StyleGAN \cite{DBLP:conf/iwbf/VenkateshZRRDB20}, MIPGAN-II \cite{DBLP:journals/corr/abs-2009-01729MIPGAN}, and our ReGenMorph. The blending artifacts in the LMA and the striping artifacts in MIPGAN-II and StyleGAN are significantly more apparent than in our proposed ReGenMorph.}
        \label{fig:samples_big}
\end{figure*}

So far, the creation of face morphing attacks has focused on the interpolation of identity information in one of two spaces. The first is on the image space, where the detected facial landmarks in both images are interpolated, and the texture information is blended \cite{DBLP:conf/icb/RaghavendraRVB17}, as demonstrated in Figure \ref{fig:pipelines}.a. The second is based on utilizing GAN structures by encoding two original images into the latent space, where the two latent vectors are interpolated (and possibly optimized \cite{DBLP:journals/corr/abs-2009-01729MIPGAN}), then the GAN generator would process the interpolated latent vector into the morphed image \cite{DBLP:conf/btas/DamerS0K18,DBLP:conf/btas/DamerBSKK19,DBLP:conf/iwbf/VenkateshZRRDB20,DBLP:journals/corr/abs-2009-01729MIPGAN}. The pipeline of creating the morphed images using GAN structures with the interpolation on the latent space is demonstrated in Figure \ref{fig:pipelines}.b, where the first approach provides very high identity preservation, the interpolation, and blending commonly results in artifacts such as shadowing (see Figure \ref{fig:samples_big}.a). These artifacts can vary between different automatic and manual morphing processes. However, they are a common feature of such morphing techniques. The second approach requires latent space manipulation that introduces irregularities in the latent vector distribution, resulting in generative artifacts in the morphed image. Such artifacts are commonly seen as striping effect and overall synthetic appearance of the image, however, with no shadowing and blending artifacts. It must be noted that existing GAN-based morphs tend to have lower identity preservation in comparison to LMA morphs. Examples of such morphs can be seen in Figures \ref{fig:samples_big}.b and \ref{fig:samples_big}.c.  

To (a) take advantage of the lower blending artifacts in the LMA morphs, (b) avoid latent space manipulation and consequent GAN artifacts in GAN morphs, and (c) make full use of the GAN capabilities in producing realistic images, we propose our ReGenMorph face morphing pipeline.

The proposed ReGenMorph pipeline is demonstrated in Figure \ref{fig:pipelines}.c, where the morphed images are first morphed identically to LMA morphs, resulting in a high-quality identity blend with blending effects. Then, the morphed image is processed into the latent space of a GAN structure, and simply re-generated by the GAN generator. This allows to take the identity blend of the LMA morph and allows the generator to create a realistic face image with minimum artifacts. This is motivated by a fact that the latent vector in this case is not manipulated and possesses distribution properties suitable for the generator (produced by its own encoder). In the following sub-section, the detailed implementation of the ReGenMorph pipeline is discussed.

\subsection{Creating ReGenMorph morphing attacks}
\label{sec:meth:ft}

Creating the ReGenMorph morphing attacks starts with the LMA morphing of face image pairs as described in \cite{DBLP:conf/icb/RaghavendraRVB17}. The LMA morph creation and database is described in more detail in section \ref{sec:exp:db}.
We use the StyleGAN architecture \cite{DBLP:conf/cvpr/KarrasLA19} as the backbone of our regeneration process. StyleGAN \cite{DBLP:conf/cvpr/KarrasLA19} is an extension to the typical GAN architecture with significant modifications to the generator model. These changes include the deployment of a mapping network to map points in latent space to an intermediate latent space, the utilization of intermediate latent space to control style at different stages in the generator model, and the introduction of noise as a variation at various stages in the generator model. This allows generating faces with varying levels of style manipulation, as well as high-quality Face images, the latter of which is interesting for the ReGenMorph pipeline.

We fine-tune the StyleGAN encoder to adapt the nature of the used images. The fine-tuned StyleGAN encoder is used to transfer the LMA morph, following face alignment and localization, into the latent space. The resulting latent vector is processed by the pre-trained StyleGAN generator to create our ReGenMorph morphing attack. This process and its technical details are presented in the following.

As a first step, the LMA morphed images are pre-processed by aligned and resized using the widely-used Multi-task Cascaded Convolutional Networks (MTCNN) \cite{zhang2016joint}. The images are resized to $1024 \times 1024$ pixels to match the StyleGAN encoder input. The training split of the data will be used to finetune the encoder of the StyleGAN.

We fine-tune the encoder of the StyleGAN to adapt to the special distribution of the LMA morphs. This is done by freezing the Generator parameters and optimizing the loss between the features extracted from the input image and the output (from the generator) image. These features are embeddings extracted using a VGG16 \cite{DBLP:journals/corr/SimonyanZ14a} network as described in \cite{DBLP:conf/cvpr/KarrasLA19}. The used loss is the mean squared error.
We used L-BFGS \cite{lbfgs} as optimizer with a learning rate of 0.25, decay rate of 0.9, early stopping with a threshold of 0.5, and patience of 10. The input image size of the StyleGAN encoder is $1024 \times 1024$ pixels, the latent vector resulting of the encoder is of the size $512 \times 1$, and the output image of the generator is a 3 channel color image of the size $1024 \times 1024$ pixels.

The process of optimizing the latent space is described in detail in \cite{DBLP:conf/icml/BojanowskiJLS18}.
The source code of the used StyleGAN along with the used pre-trained models are available\footnote{https://github.com/Puzer/stylegan-encoder} as open-source. The pre-trained StyleGAN was trained on the FFHQ \cite{DBLP:conf/cvpr/KarrasLA19} and CelebA-HQ databases \cite{DBLP:conf/iclr/KarrasALL18}, while the pre-training hyperparameters are listed in \cite{DBLP:conf/cvpr/KarrasLA19}.

To generate a ReGenMorph attack image, we create the LMA attack, detect, align, and resize the face as mentioned earlier. This attack is passed to the encoder and then the generator of the fine-tuned StyleGAN. The resulting image is our ReGenMorph. We generate a complete set of the training and testing data splits (described in Section \ref{sec:exp:db}). The training set is the same as the one used for fine-tuning and will only be used to train the MAD. All the examples, vulnerability analyses, and detectability results in this work are performed on the testing set. More details on the data structure are presented in Section \ref{sec:exp:db}.

\begin{figure*}[h!]
     \centering
     \begin{subfigure}[b]{0.09\textwidth}
         \centering
         \includegraphics[height=2.0\textwidth]{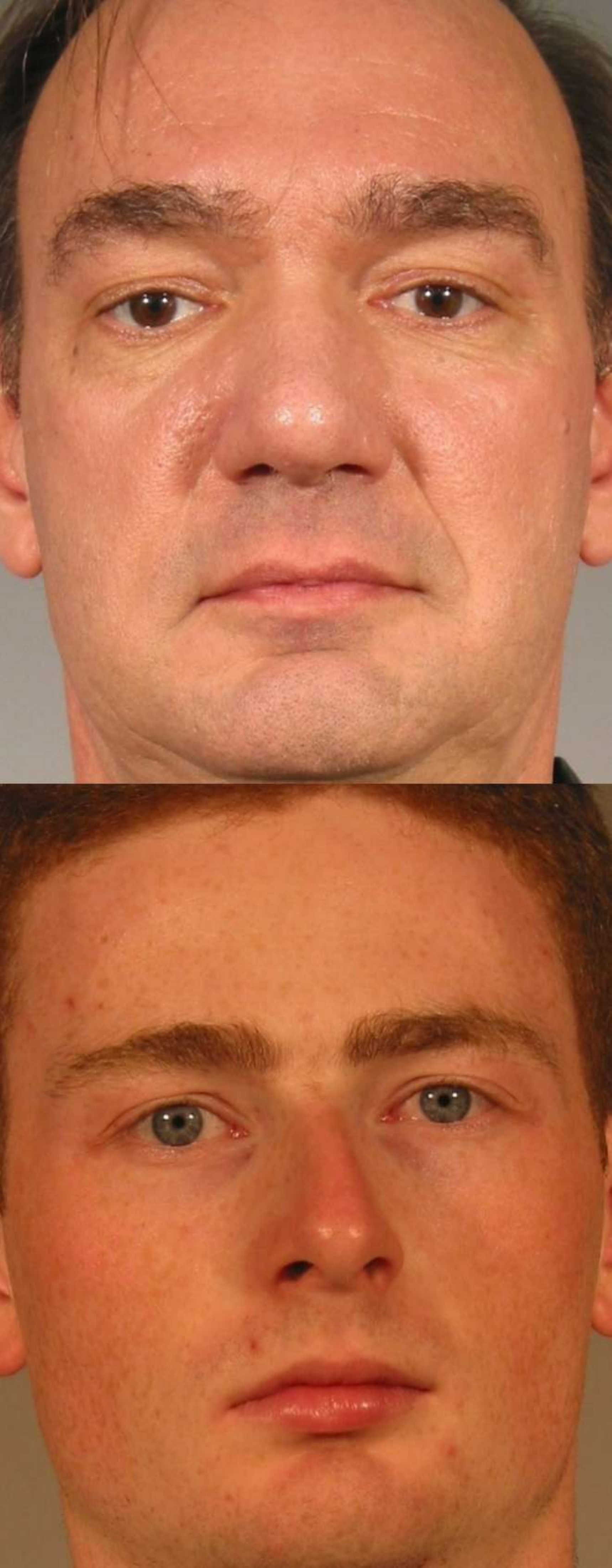}
         \\
         \includegraphics[height=2.0\textwidth]{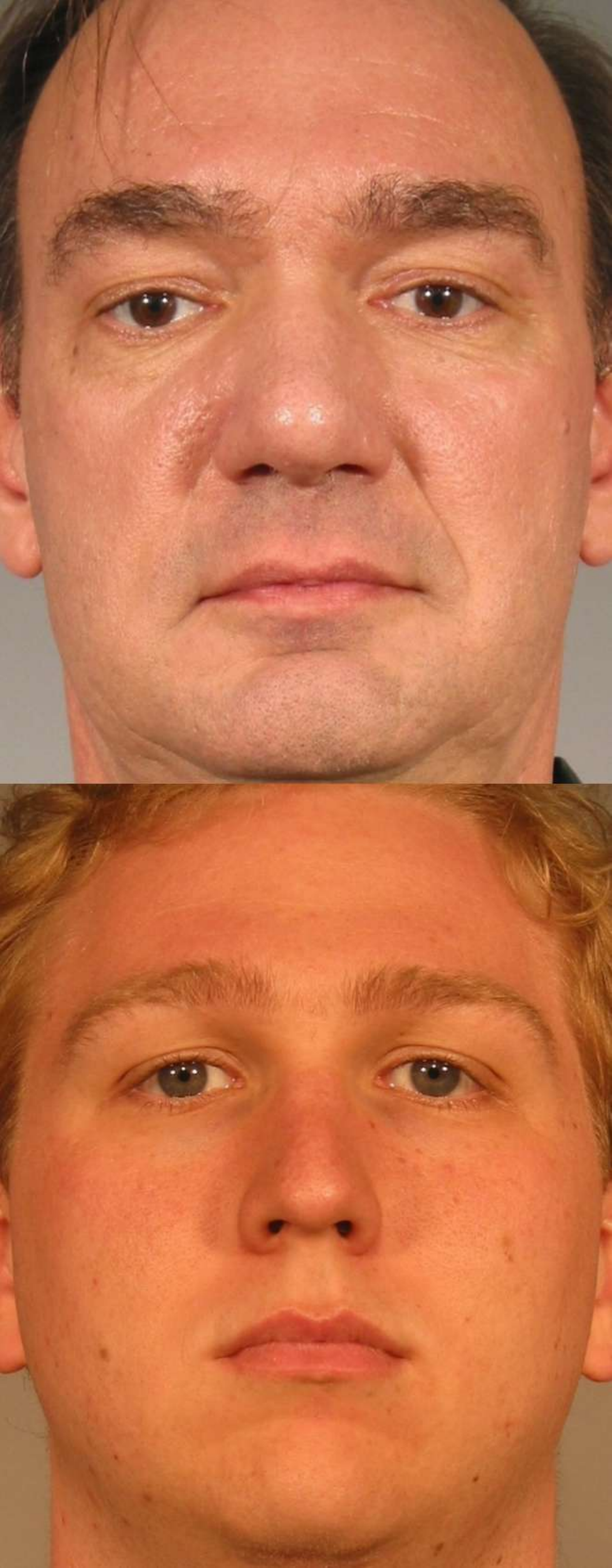}
         \\
         \includegraphics[height=2.0\textwidth]{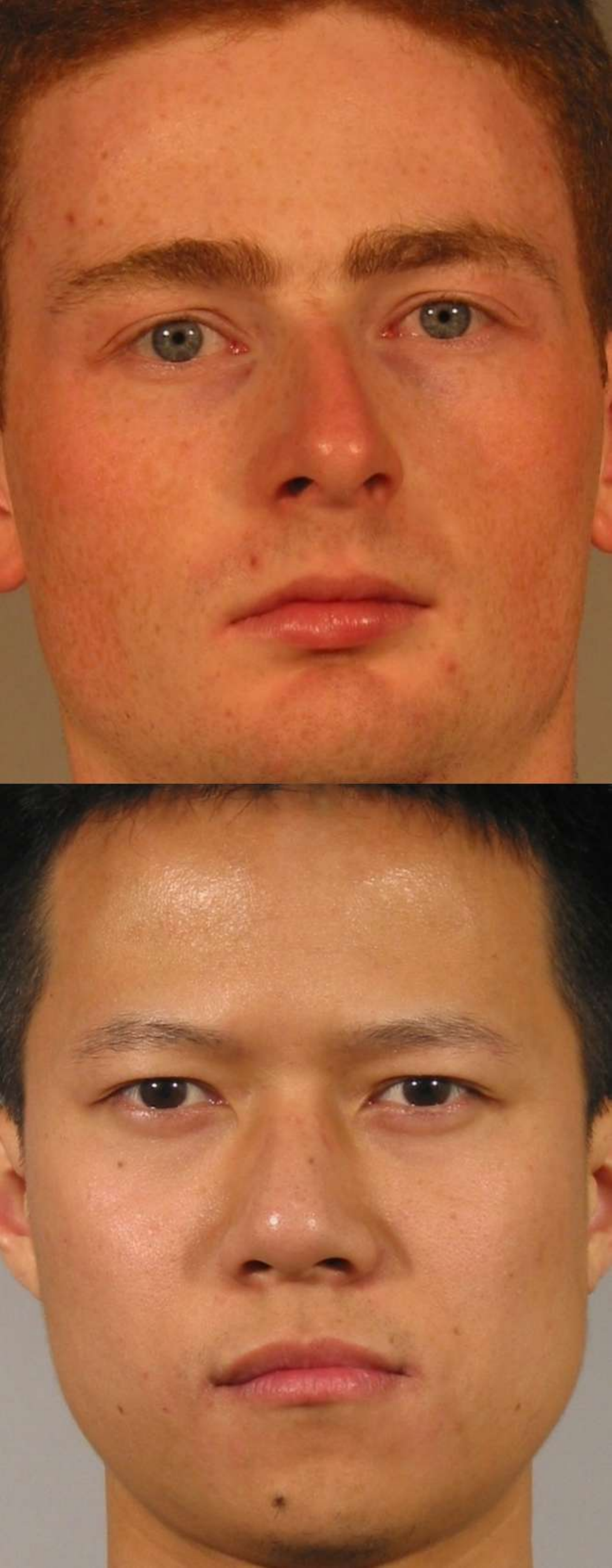}
         \\
         \includegraphics[height=2.0\textwidth]{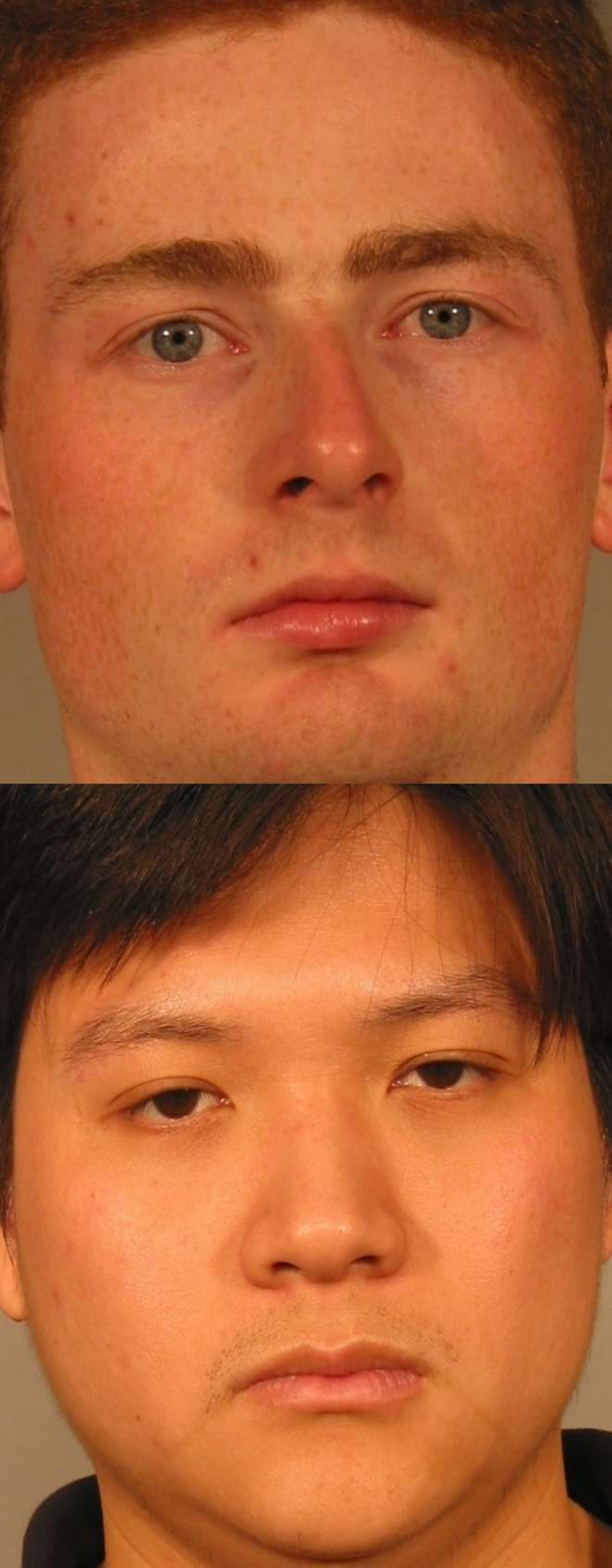}
         \\
         \includegraphics[height=2.0\textwidth]{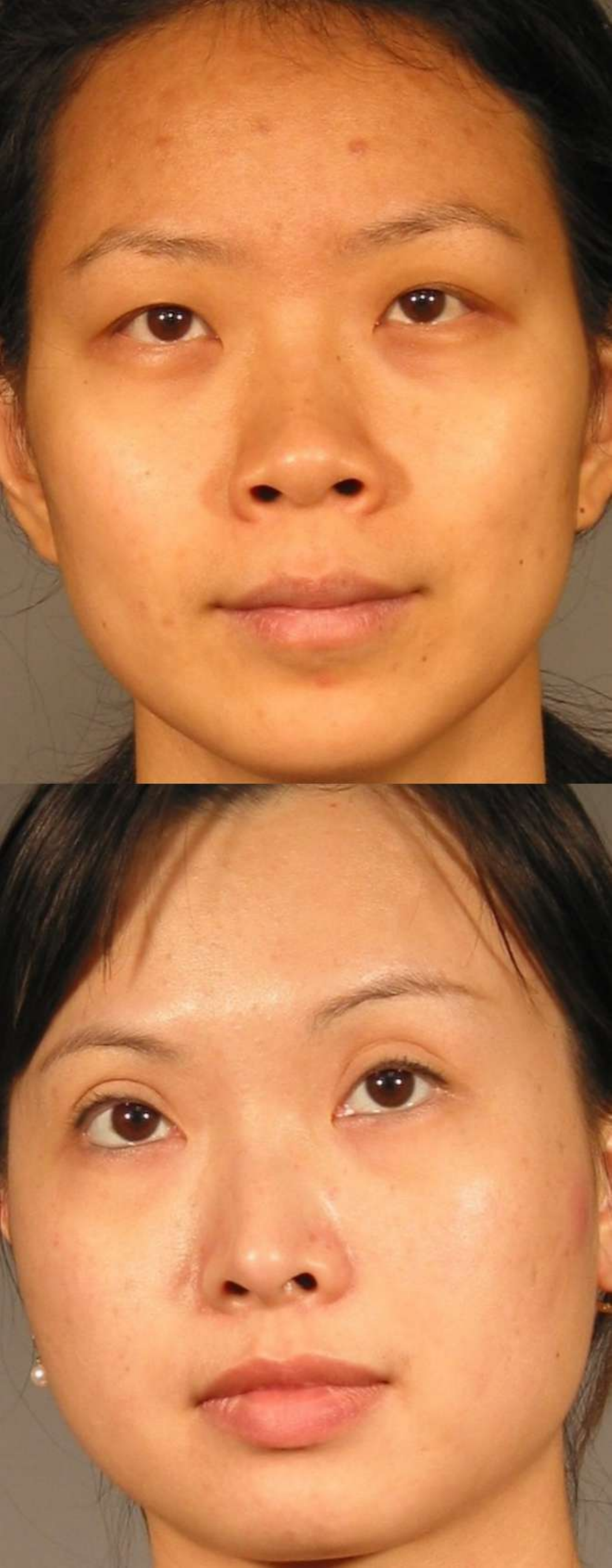}
         \\
         \includegraphics[height=2.0\textwidth]{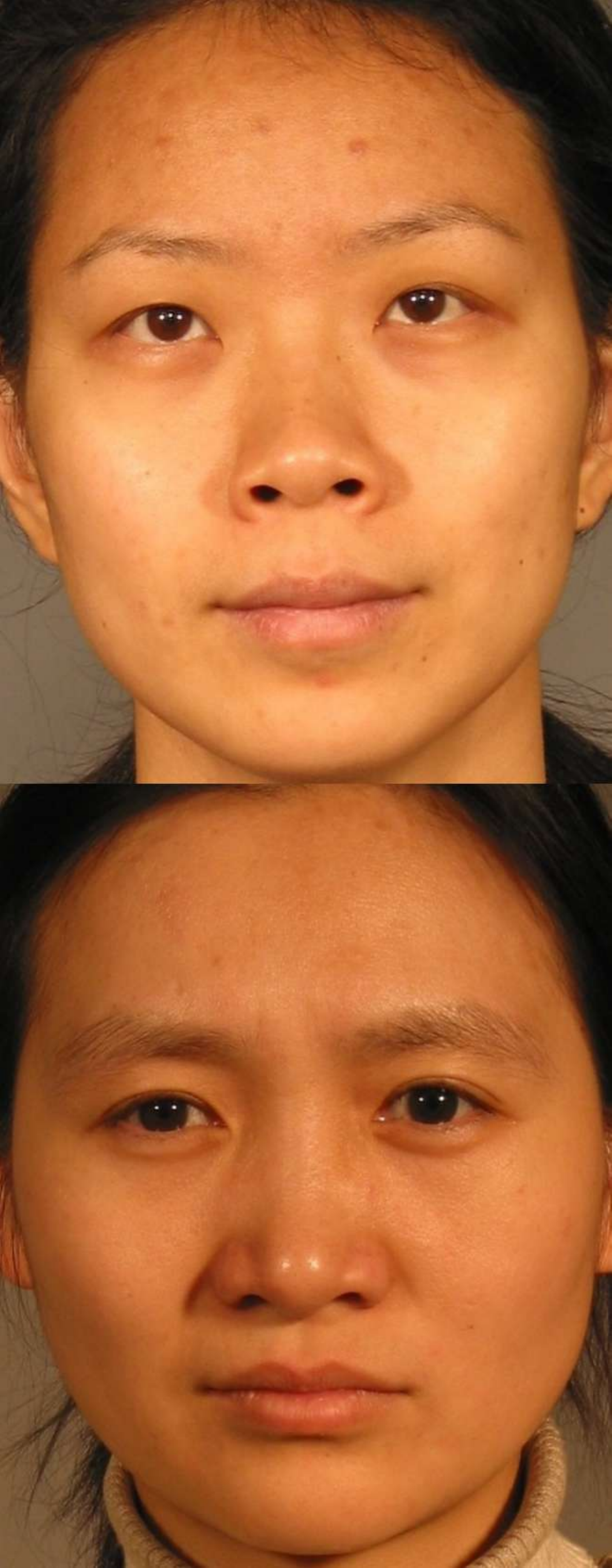}
         \caption{\tiny{BF}}
        \label{fig:samp:LMA}
     \end{subfigure}
     \begin{subfigure}[b]{0.18\textwidth}
         \centering
         \includegraphics[height=\textwidth]{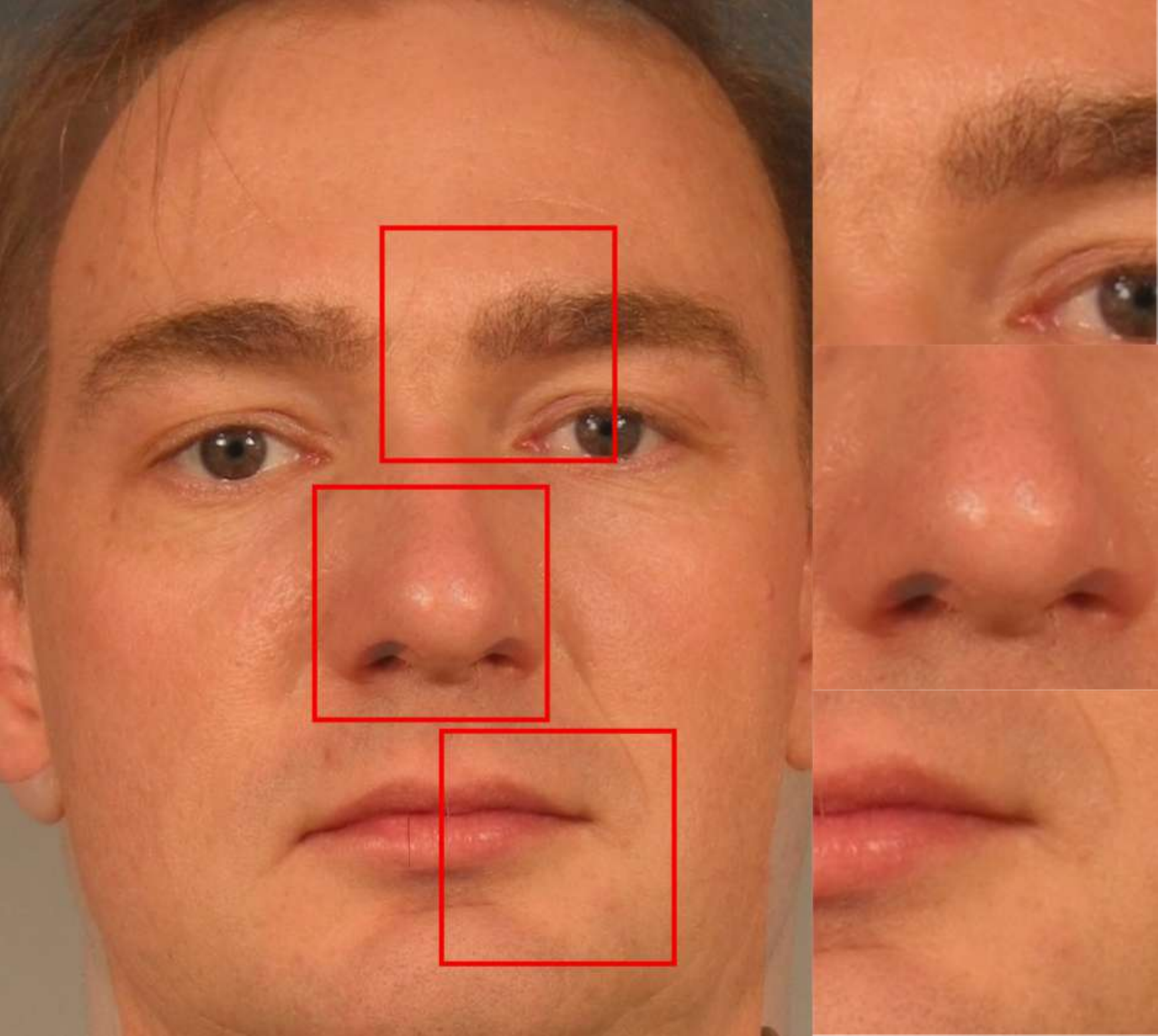}
         \\
         \includegraphics[height=\textwidth]{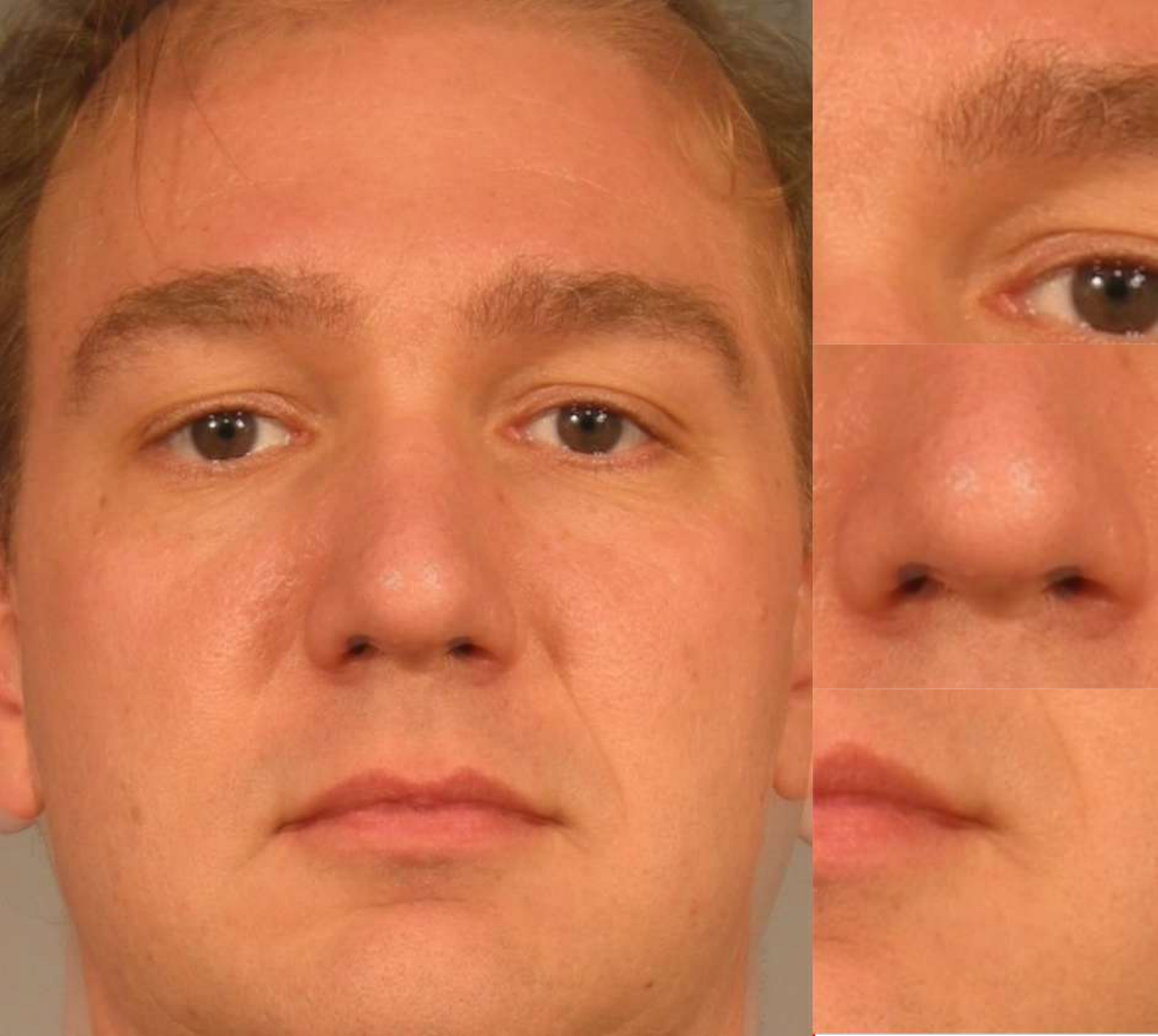}
         \\
         \includegraphics[height=\textwidth]{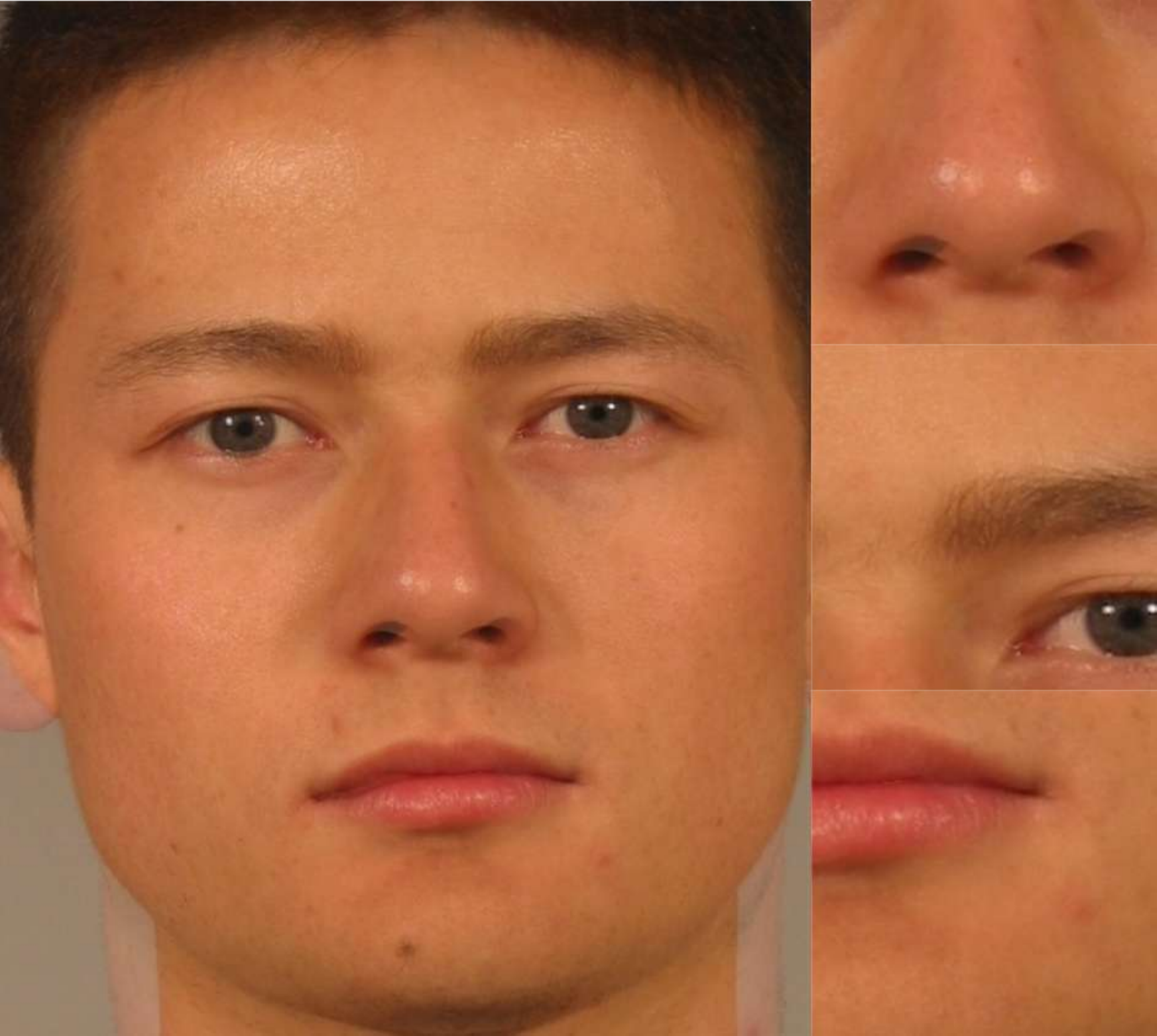}
         \\
         \includegraphics[height=\textwidth]{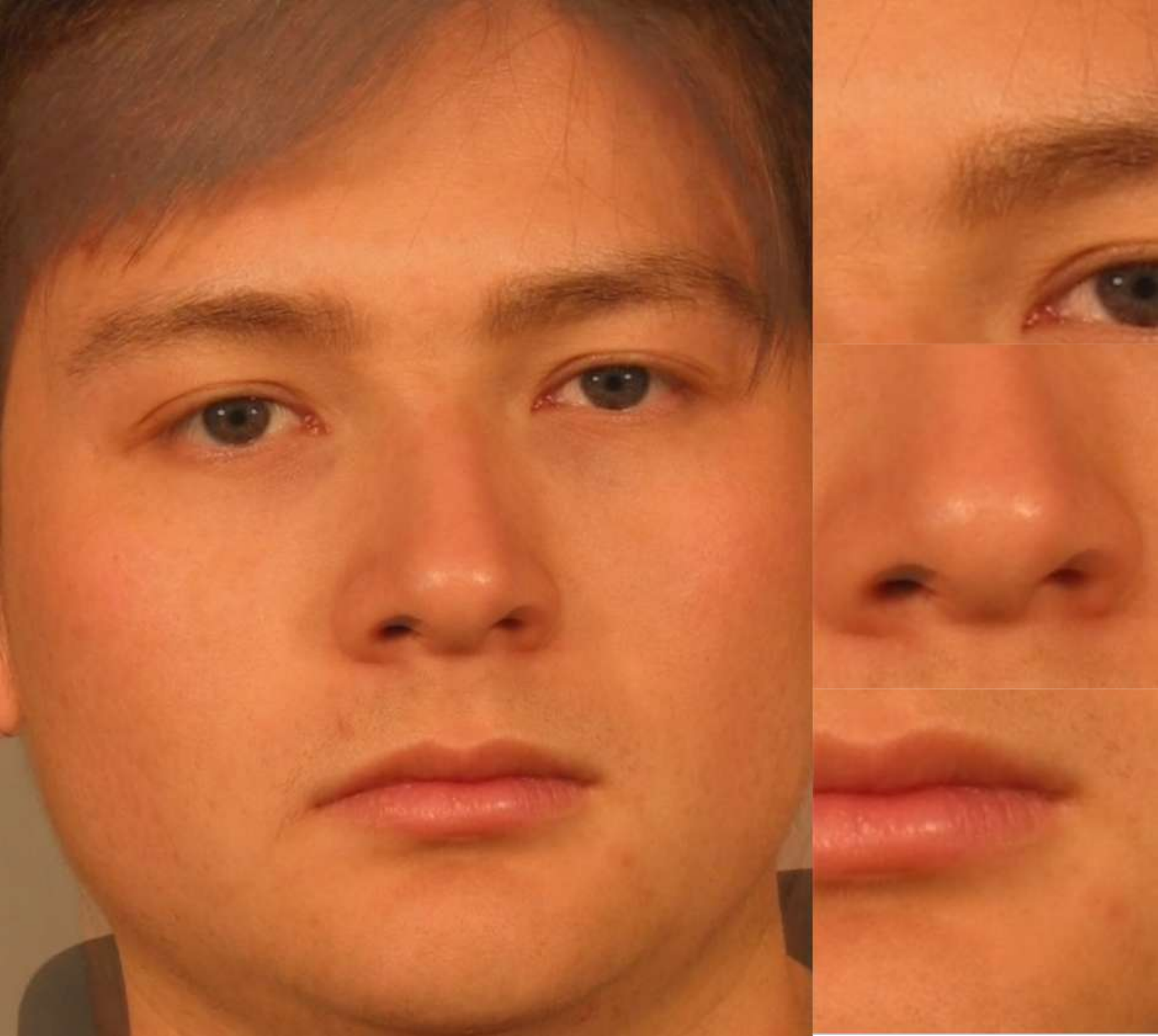}
         \\
         \includegraphics[height=\textwidth]{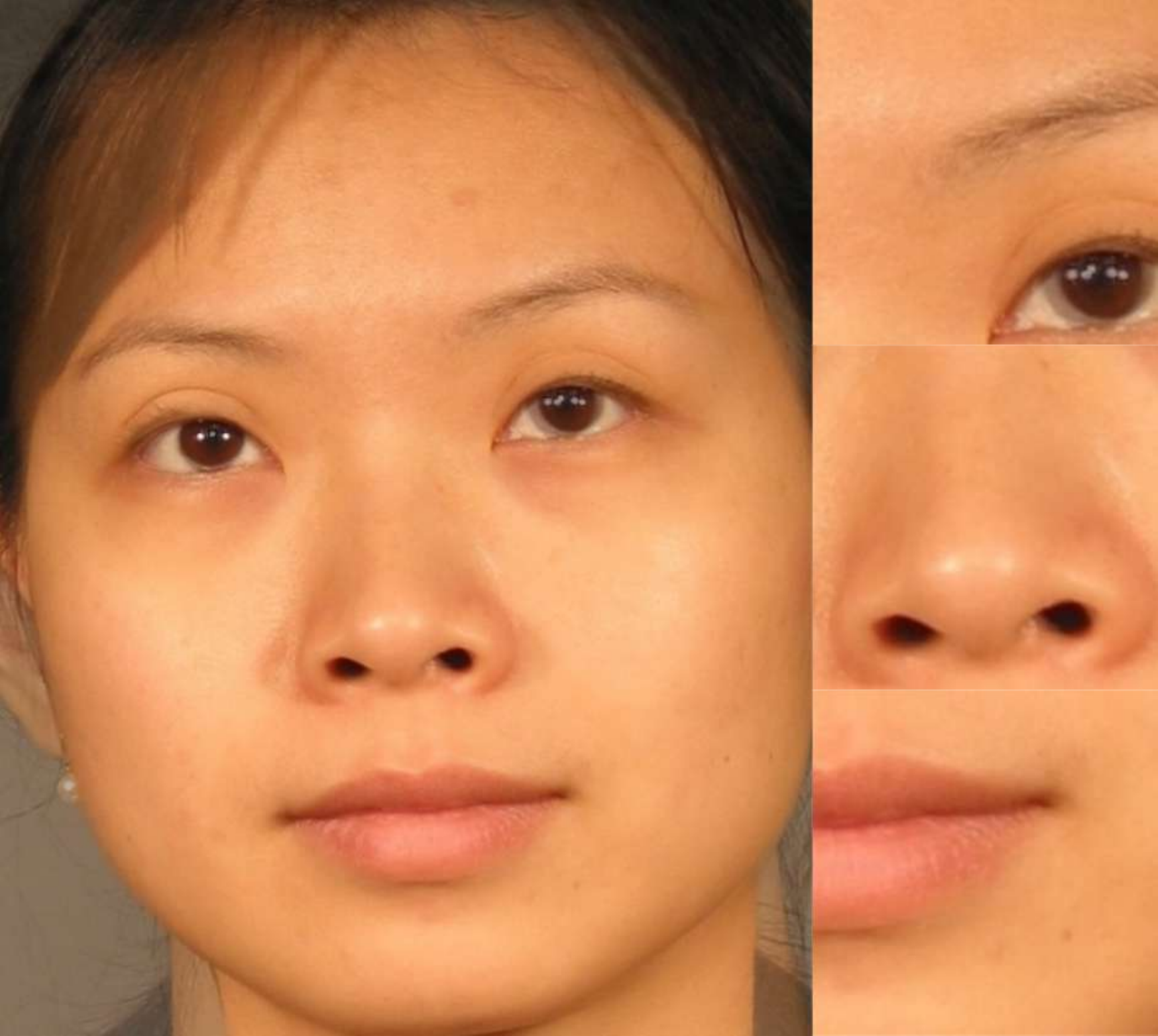}
         \\
         \includegraphics[height=\textwidth]{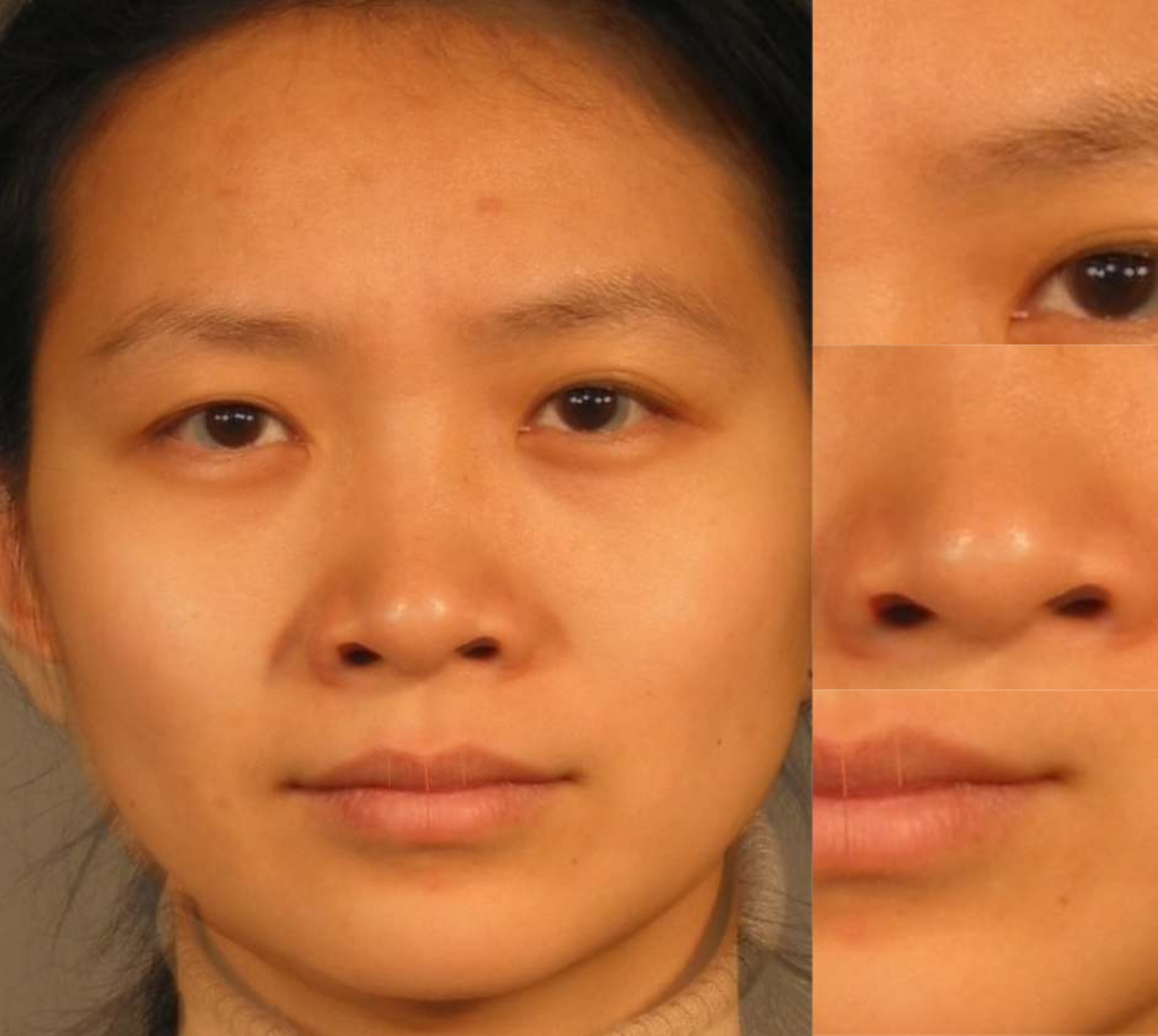}
         \caption{\tiny{LMA \cite{DBLP:conf/icb/RaghavendraRVB17}}}
        \label{fig:samp:LMA}
     \end{subfigure}
     \hfill
     \begin{subfigure}[b]{0.18\textwidth}
         \centering
         \includegraphics[height=\textwidth]{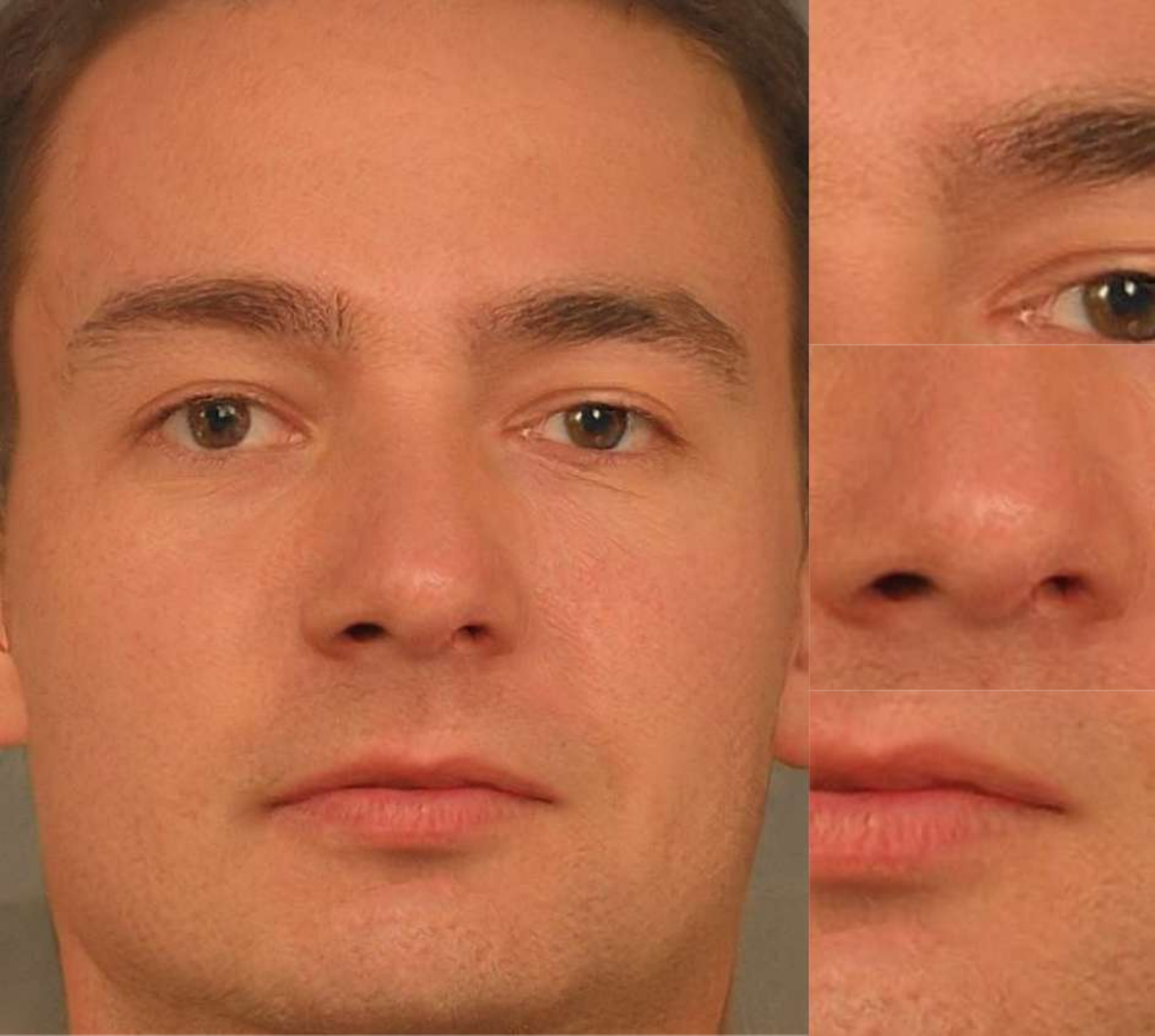}
         \\
         \includegraphics[height=\textwidth]{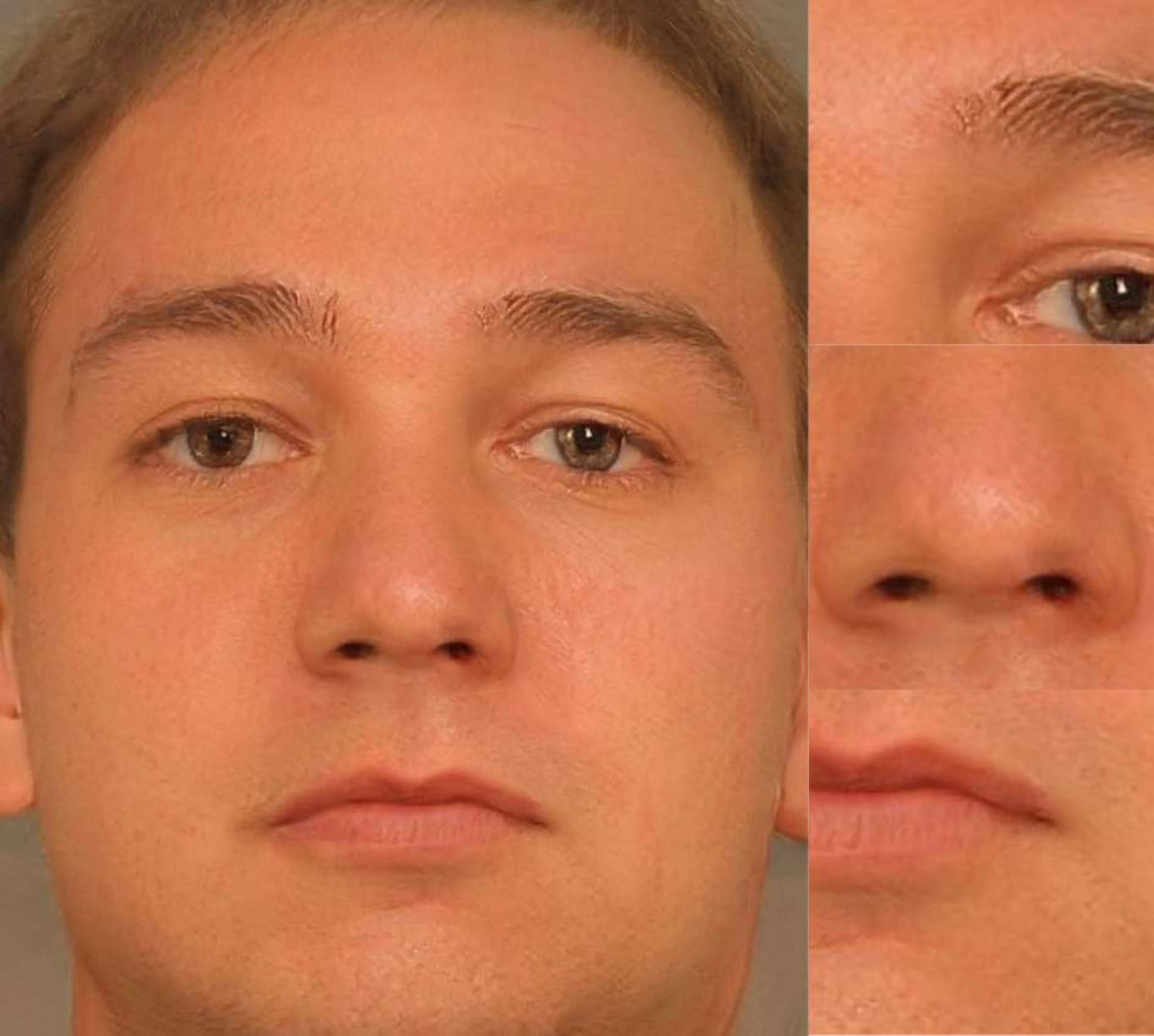}
         \\
         \includegraphics[height=\textwidth]{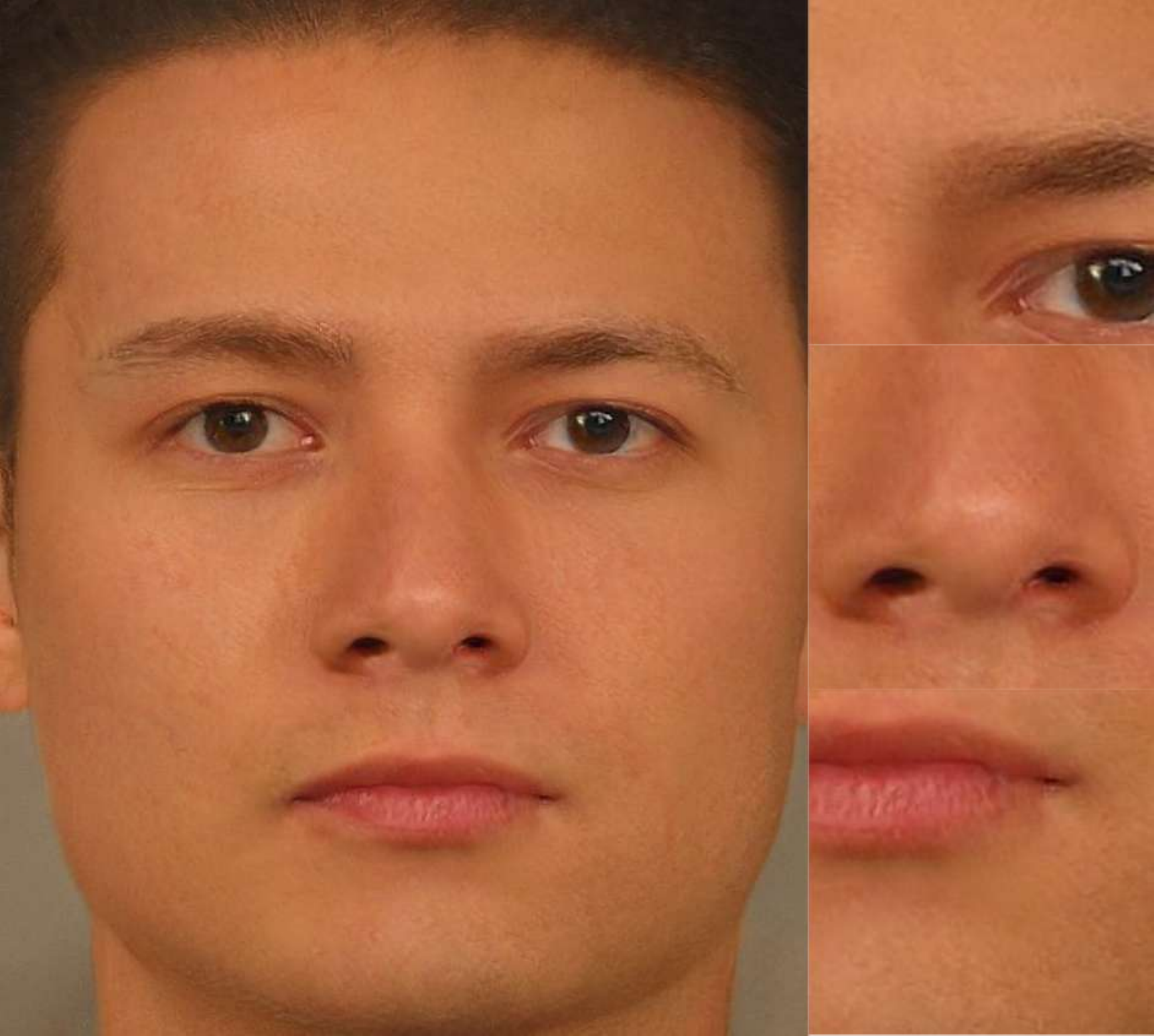}
         \\
         \includegraphics[height=\textwidth]{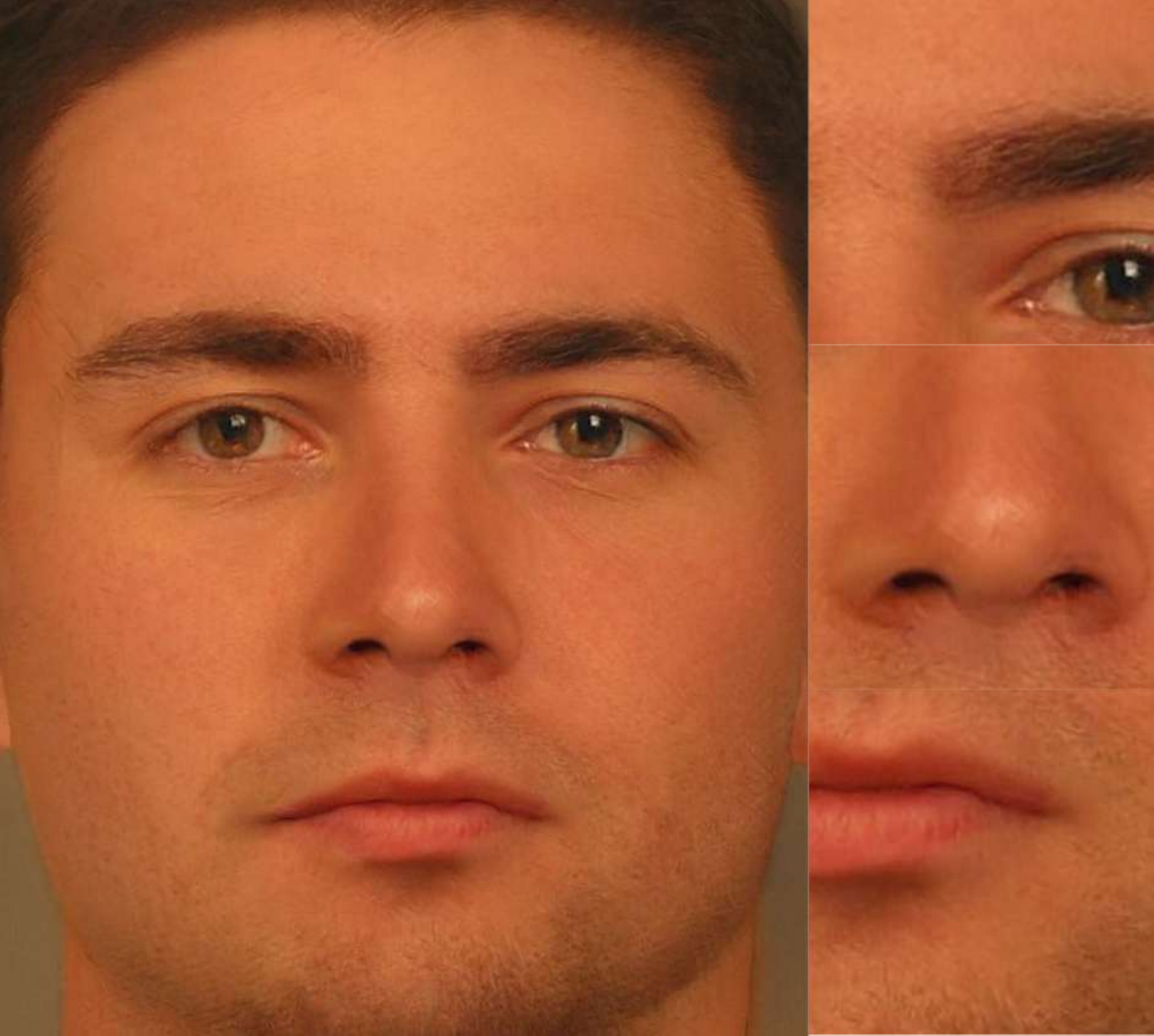}
          \\
         \includegraphics[height=\textwidth]{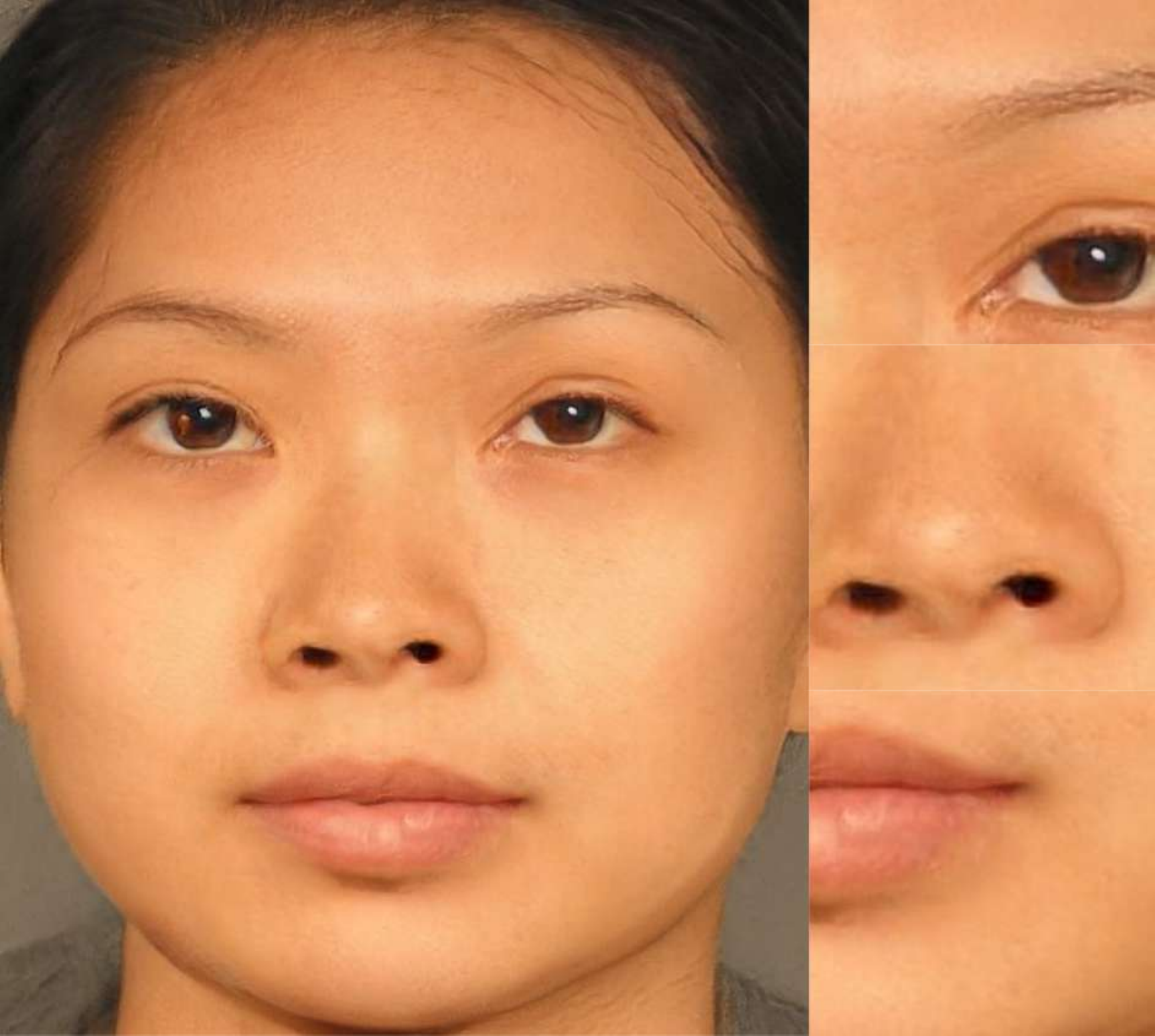}
         \\
         \includegraphics[height=\textwidth]{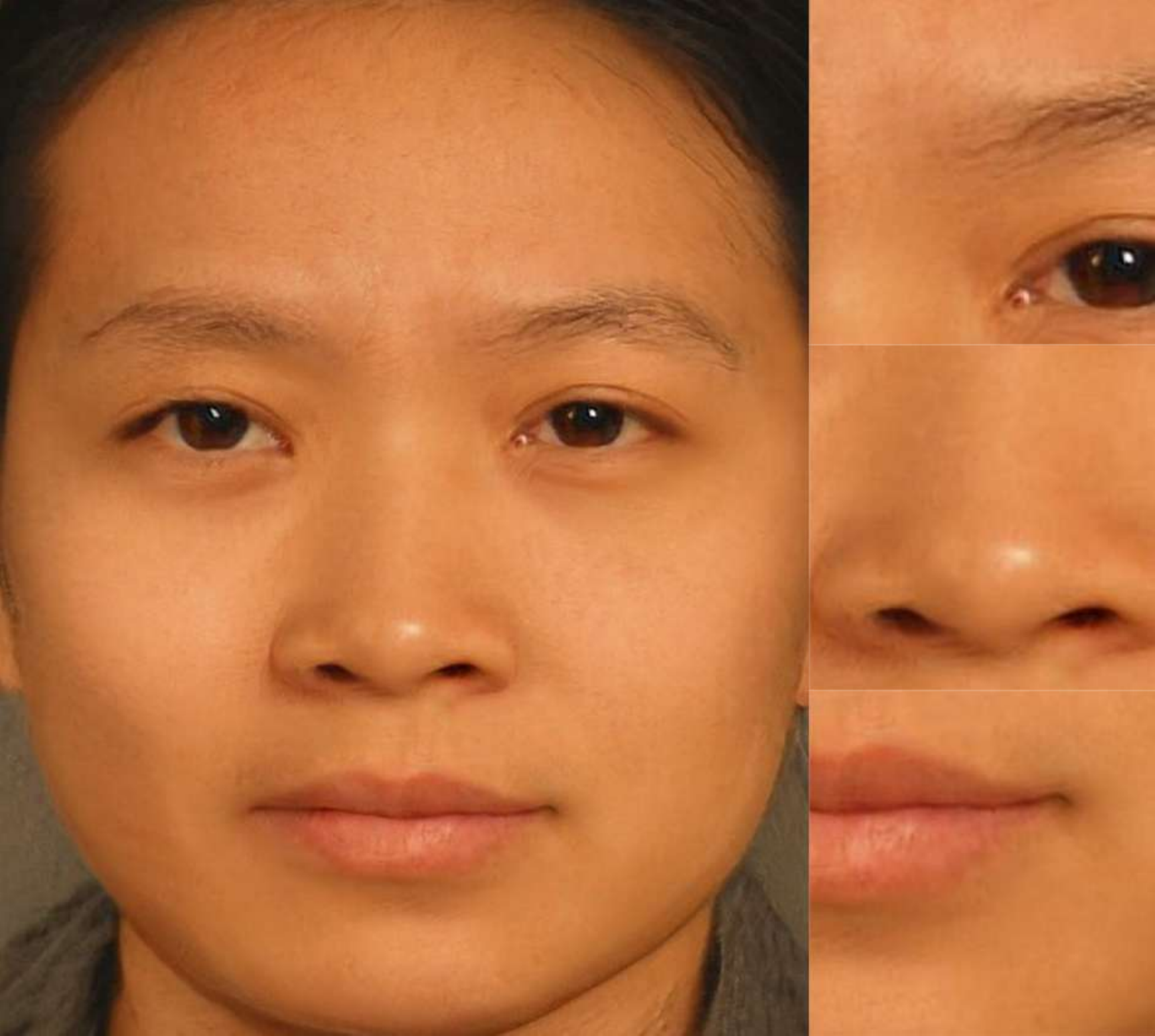}
         \caption{\tiny{StyleGAN \cite{DBLP:conf/iwbf/VenkateshZRRDB20}}}
         \label{fig:samp:BL}
     \end{subfigure}
     \hfill
     \begin{subfigure}[b]{0.18\textwidth}
         \centering
         \includegraphics[height=\textwidth]{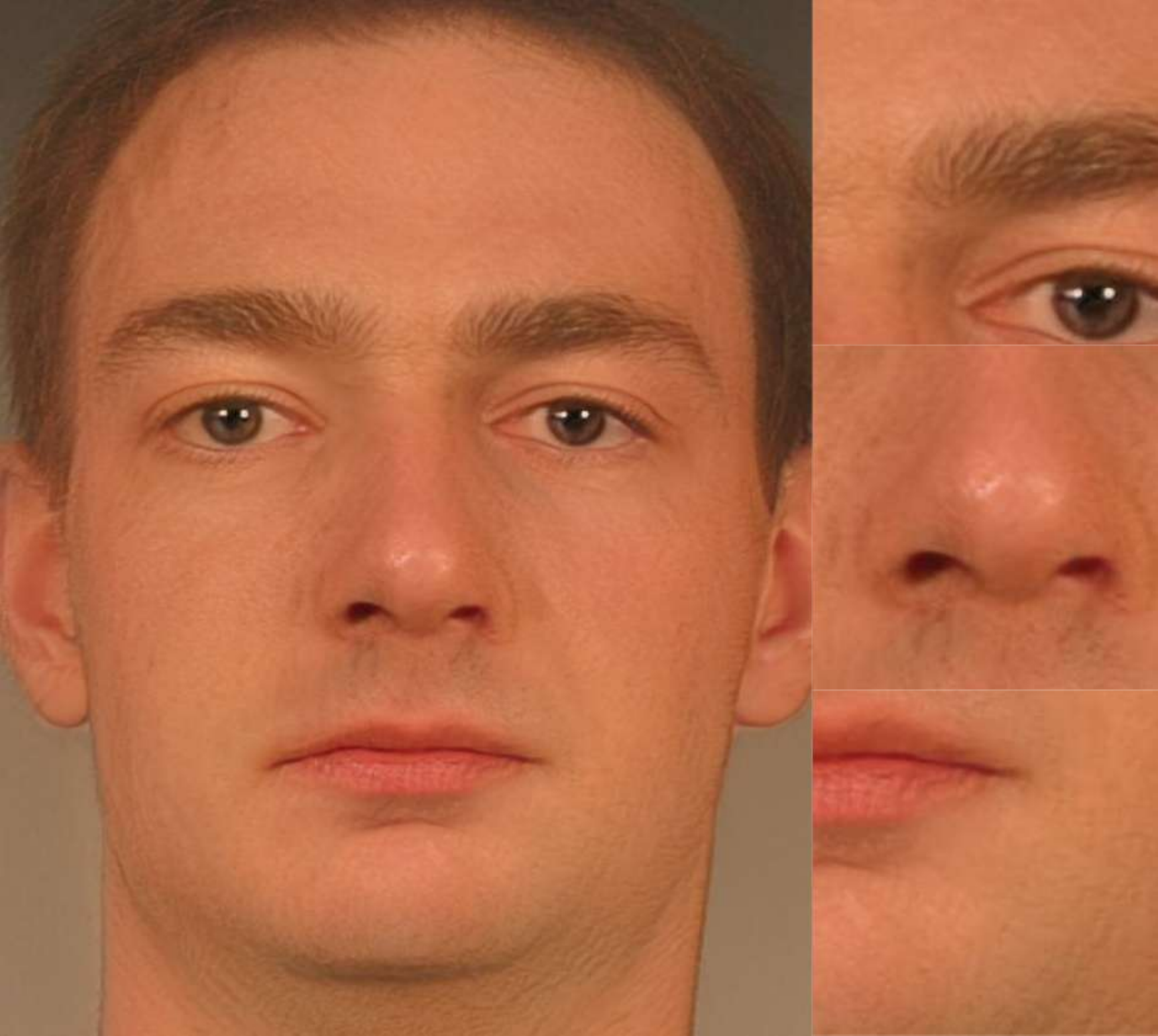}
         \\
         \includegraphics[height=\textwidth]{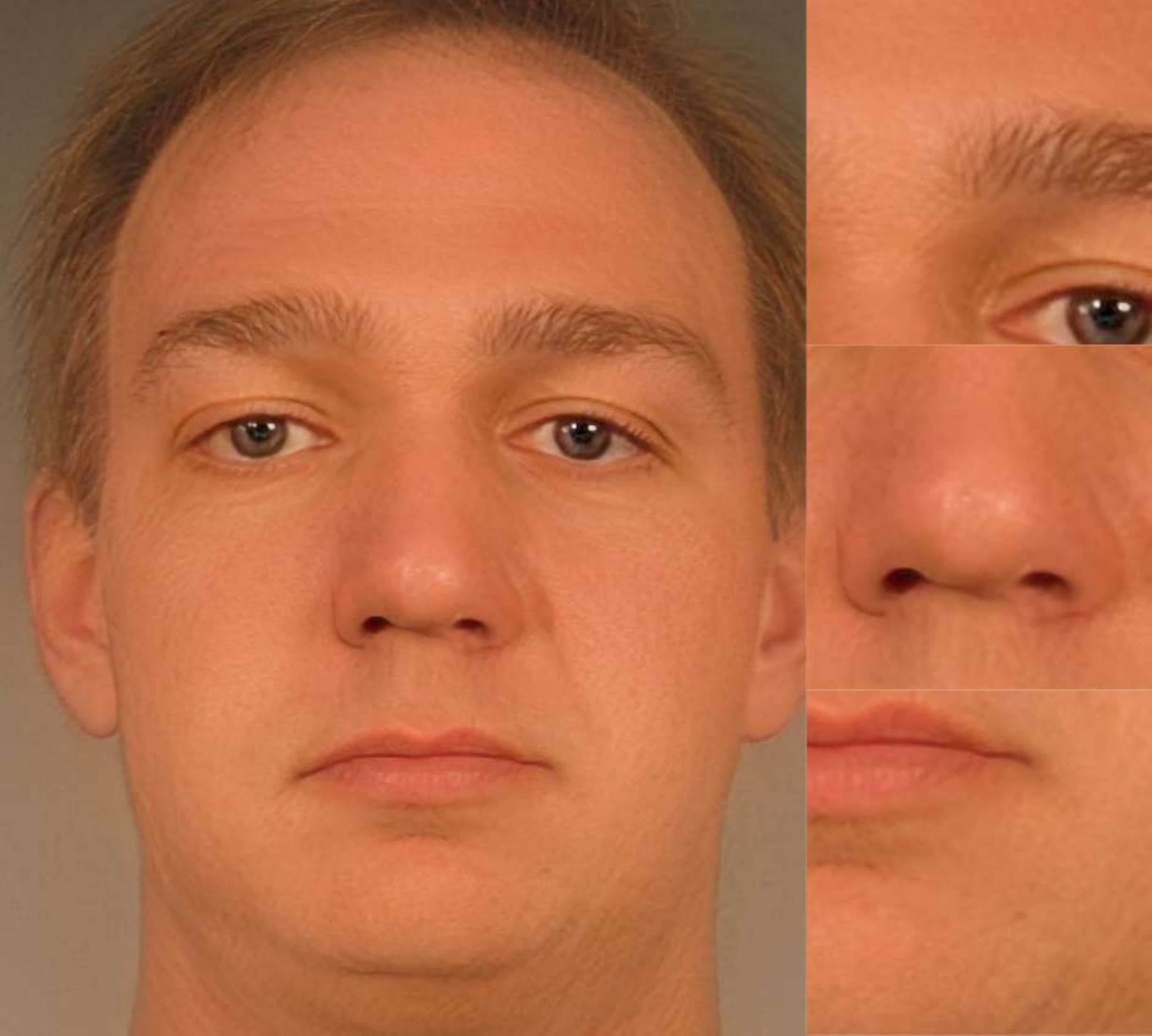}
         \\
         \includegraphics[height=\textwidth]{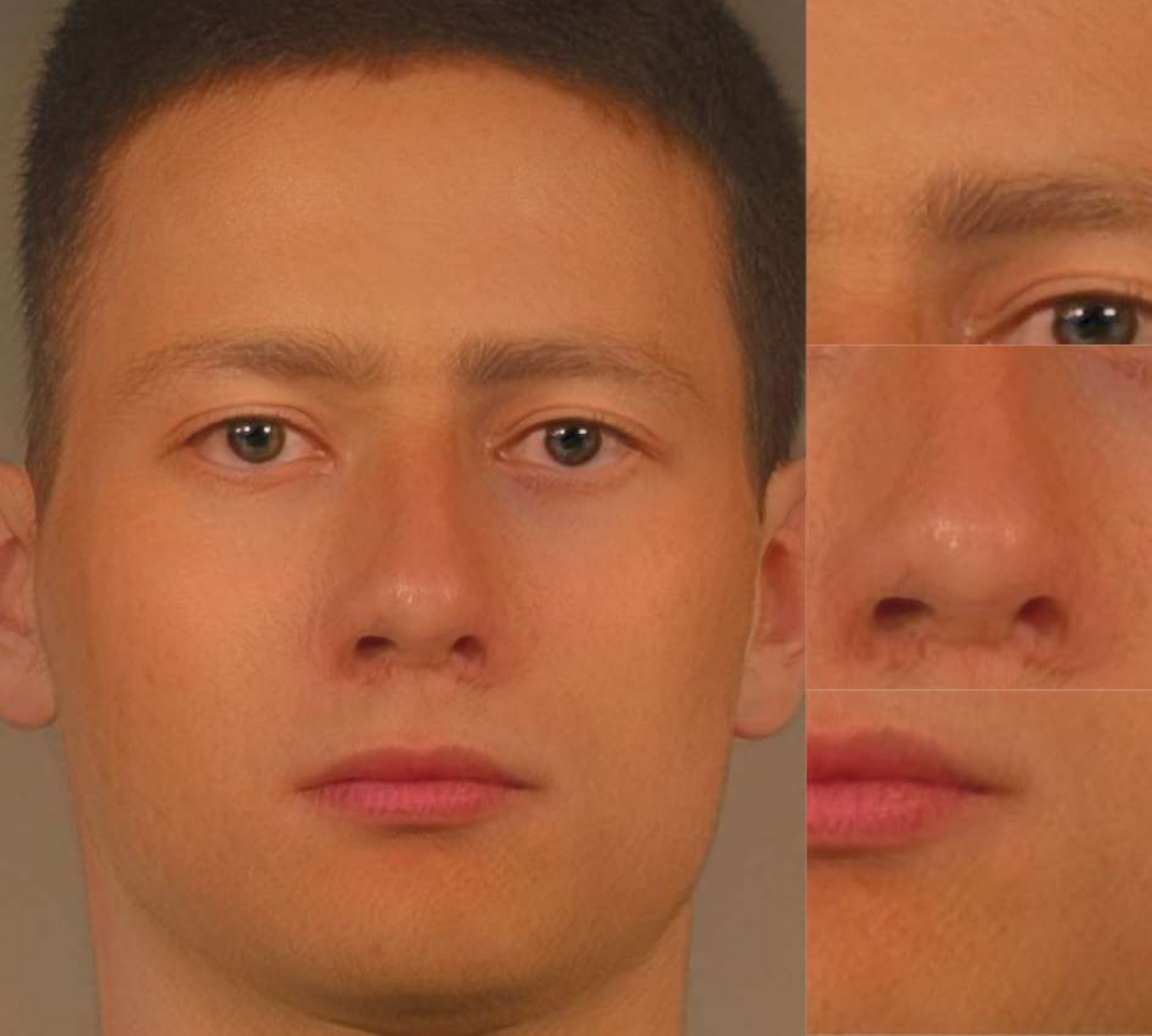}
          \\
         \includegraphics[height=\textwidth]{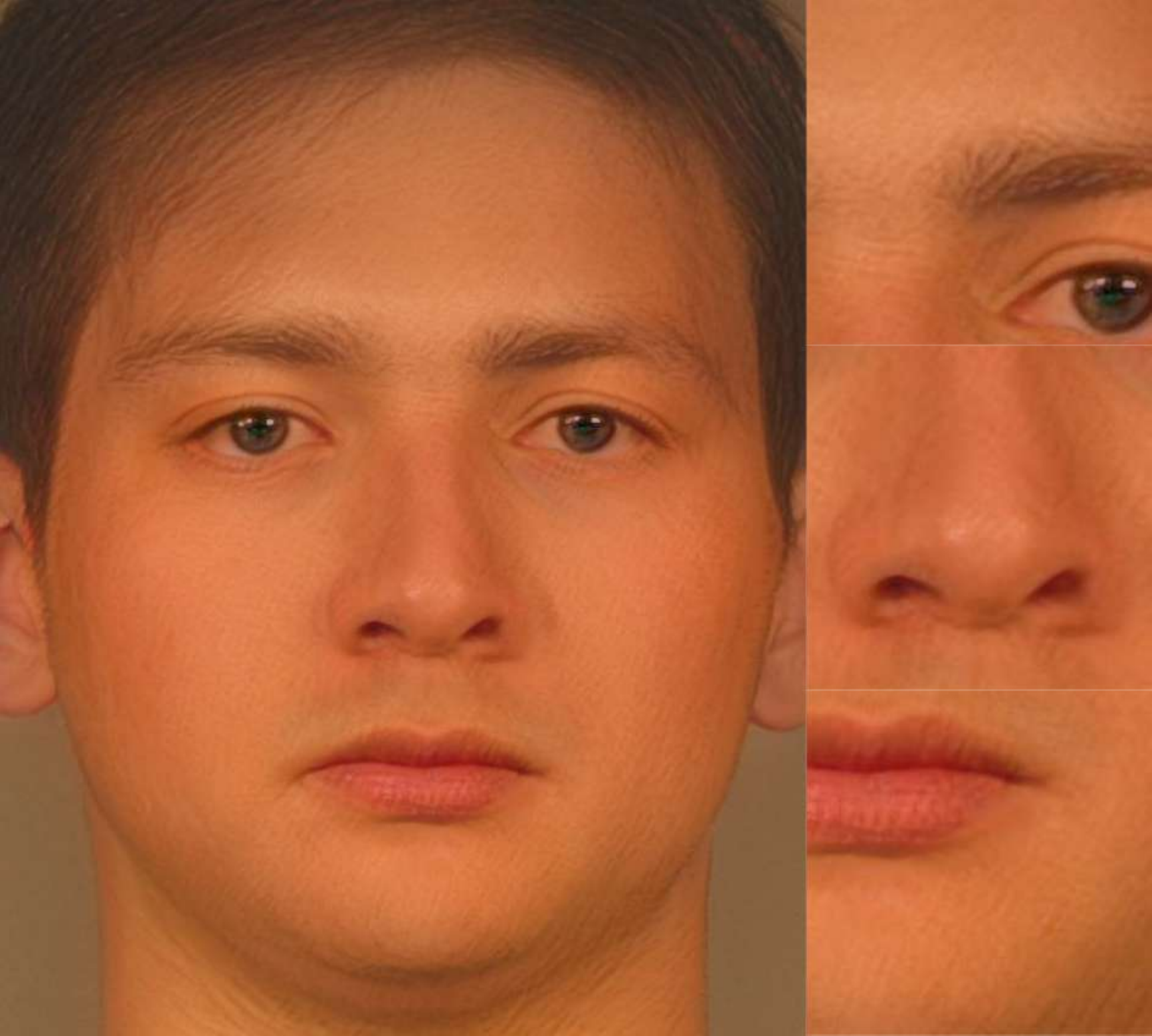}
         \\
         \includegraphics[height=\textwidth]{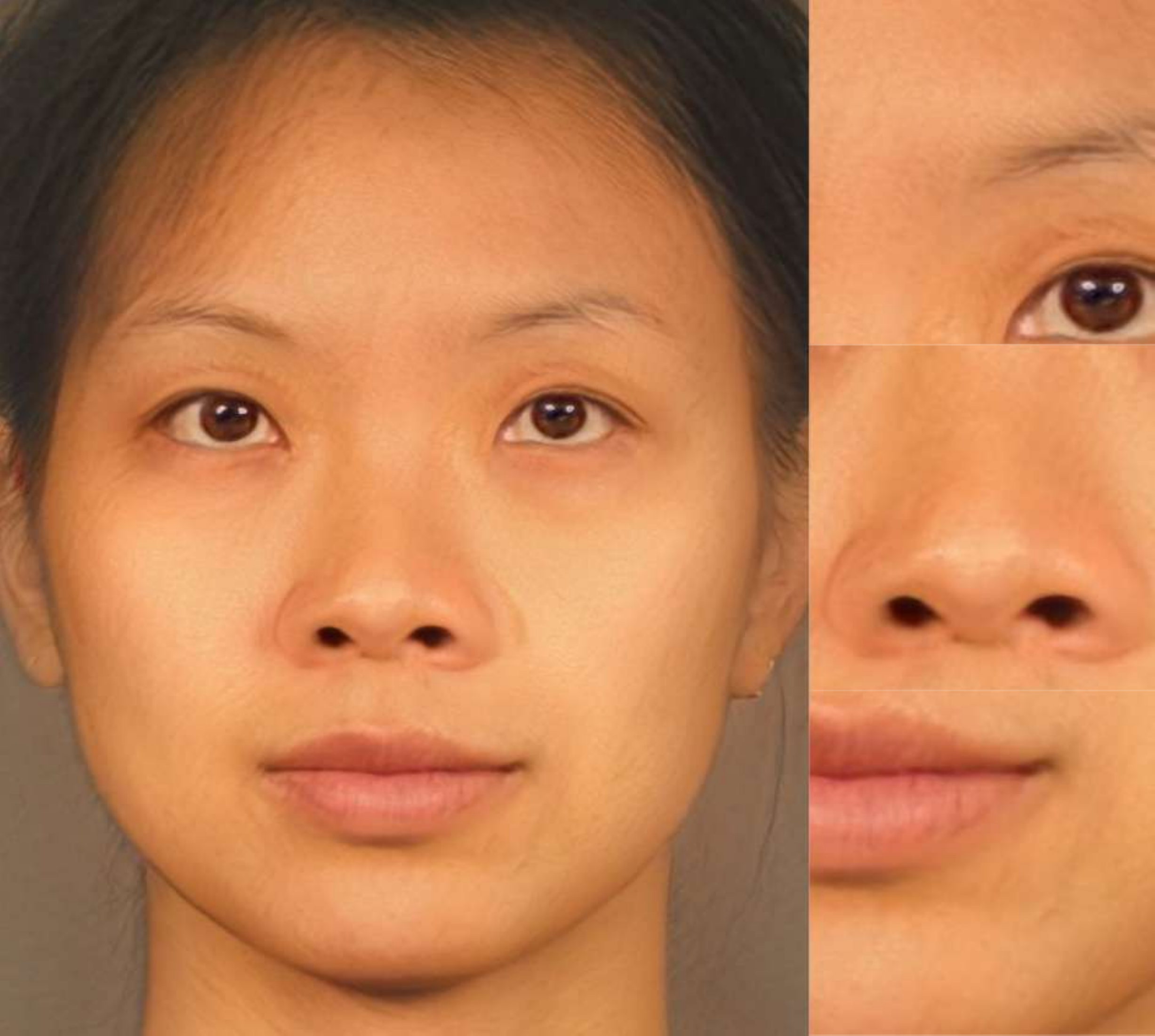}
         \\
         \includegraphics[height=\textwidth]{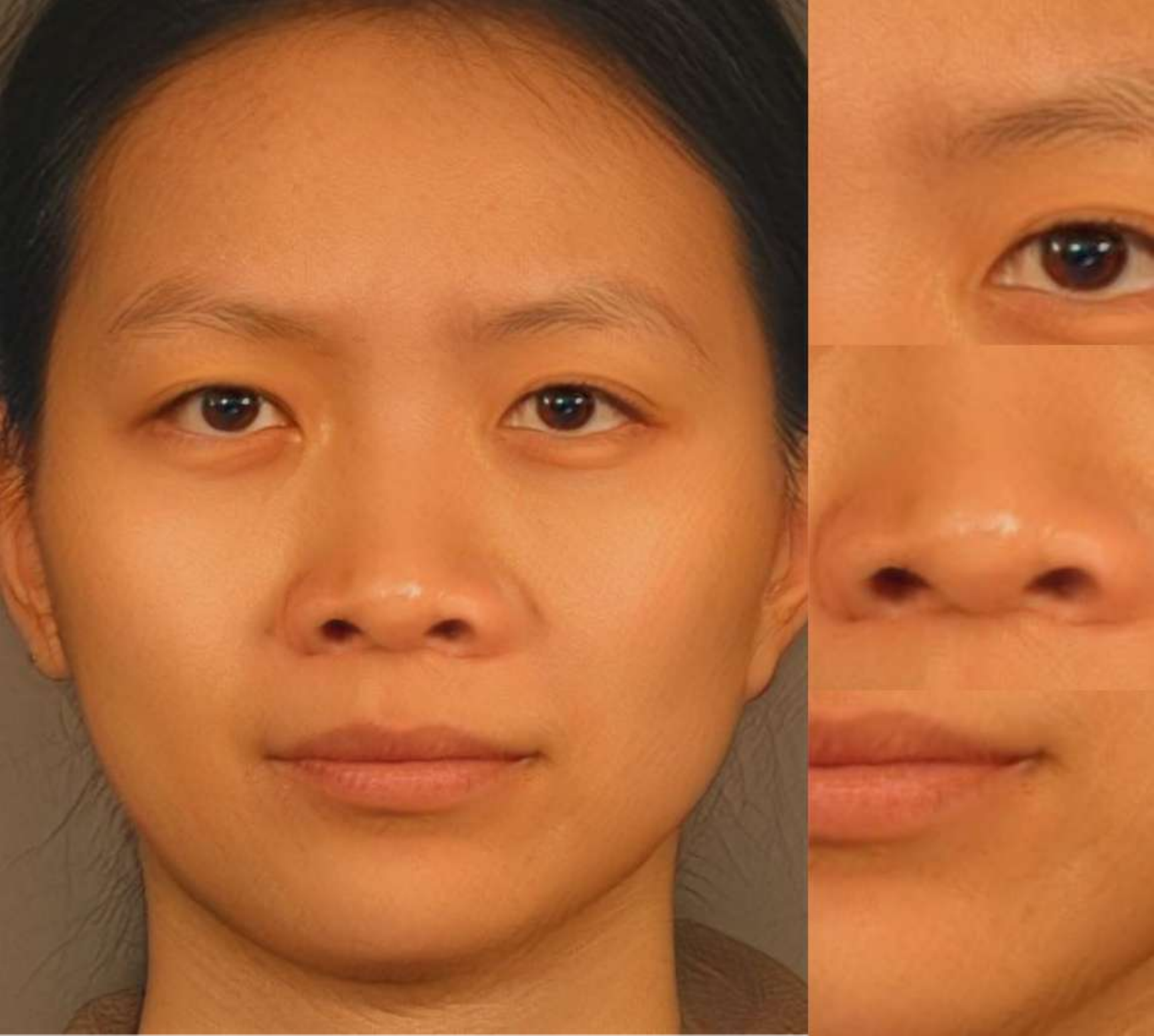}
         \caption{\tiny{MIPGAN-II \cite{DBLP:journals/corr/abs-2009-01729MIPGAN}}}
         \label{fig:samp:BL}
     \end{subfigure}
     \hfill
     \begin{subfigure}[b]{0.18\textwidth}
         \centering
         \includegraphics[height=\textwidth]{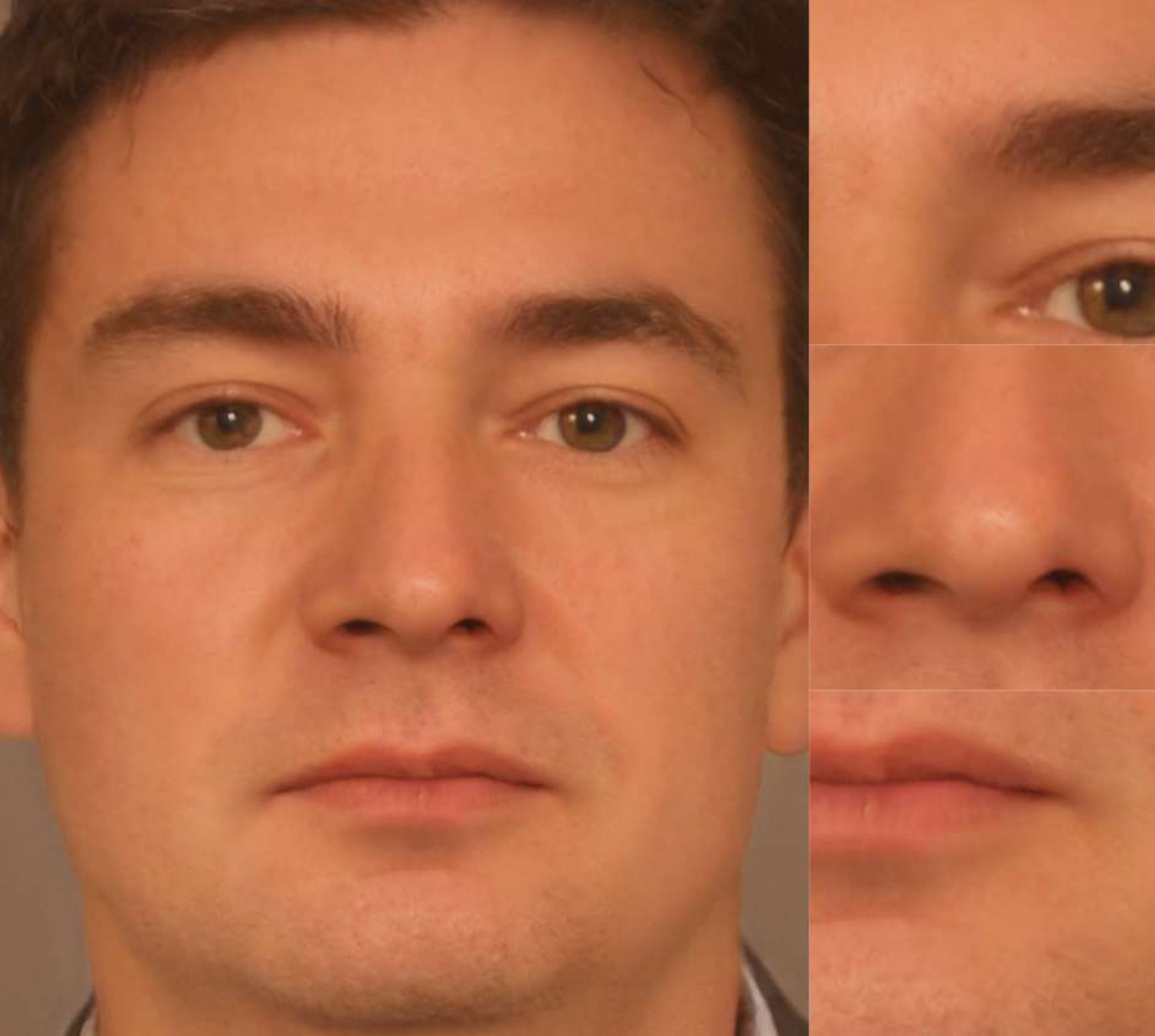}
         \\
         \includegraphics[height=\textwidth]{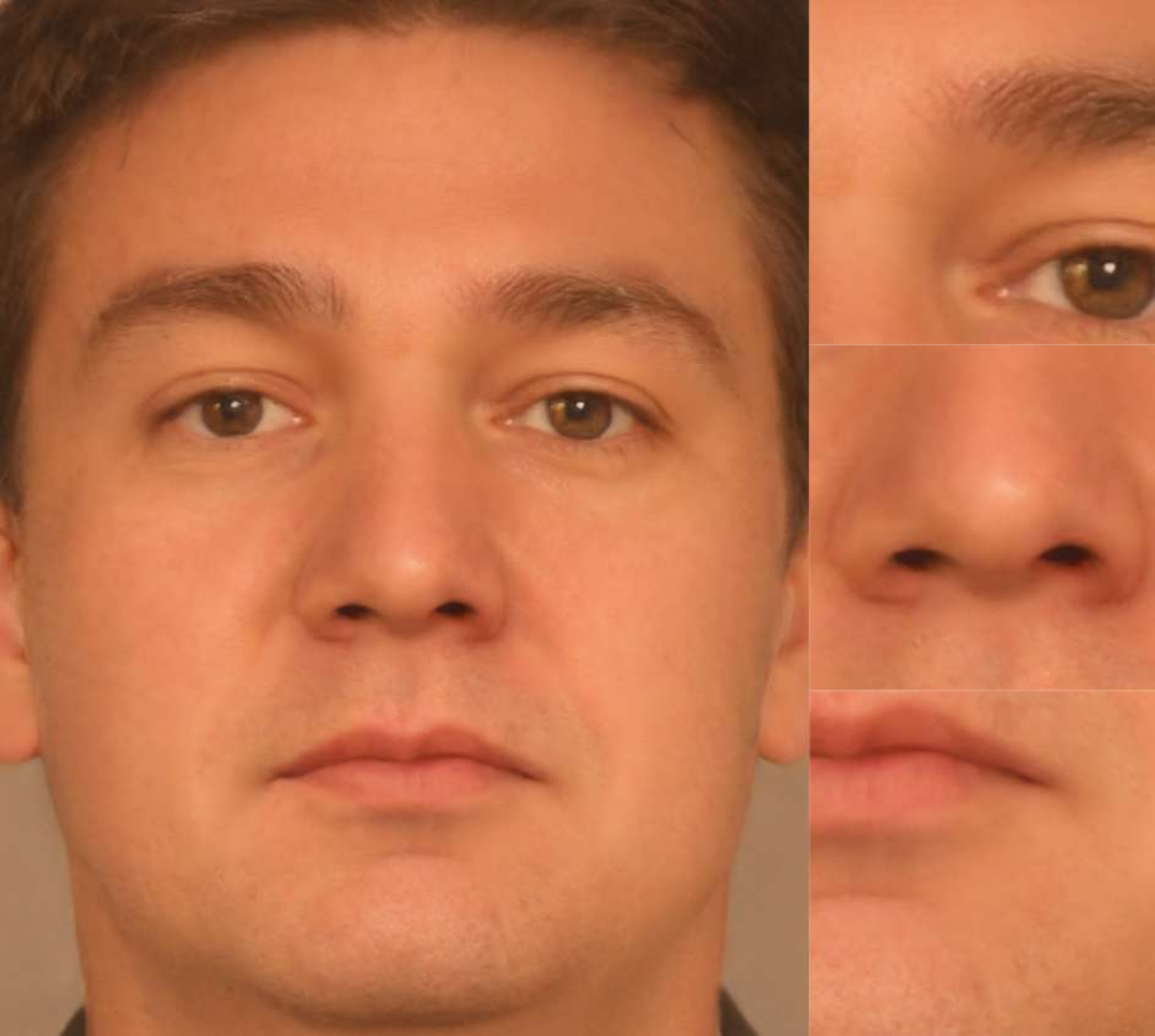}
         \\
         \includegraphics[height=\textwidth]{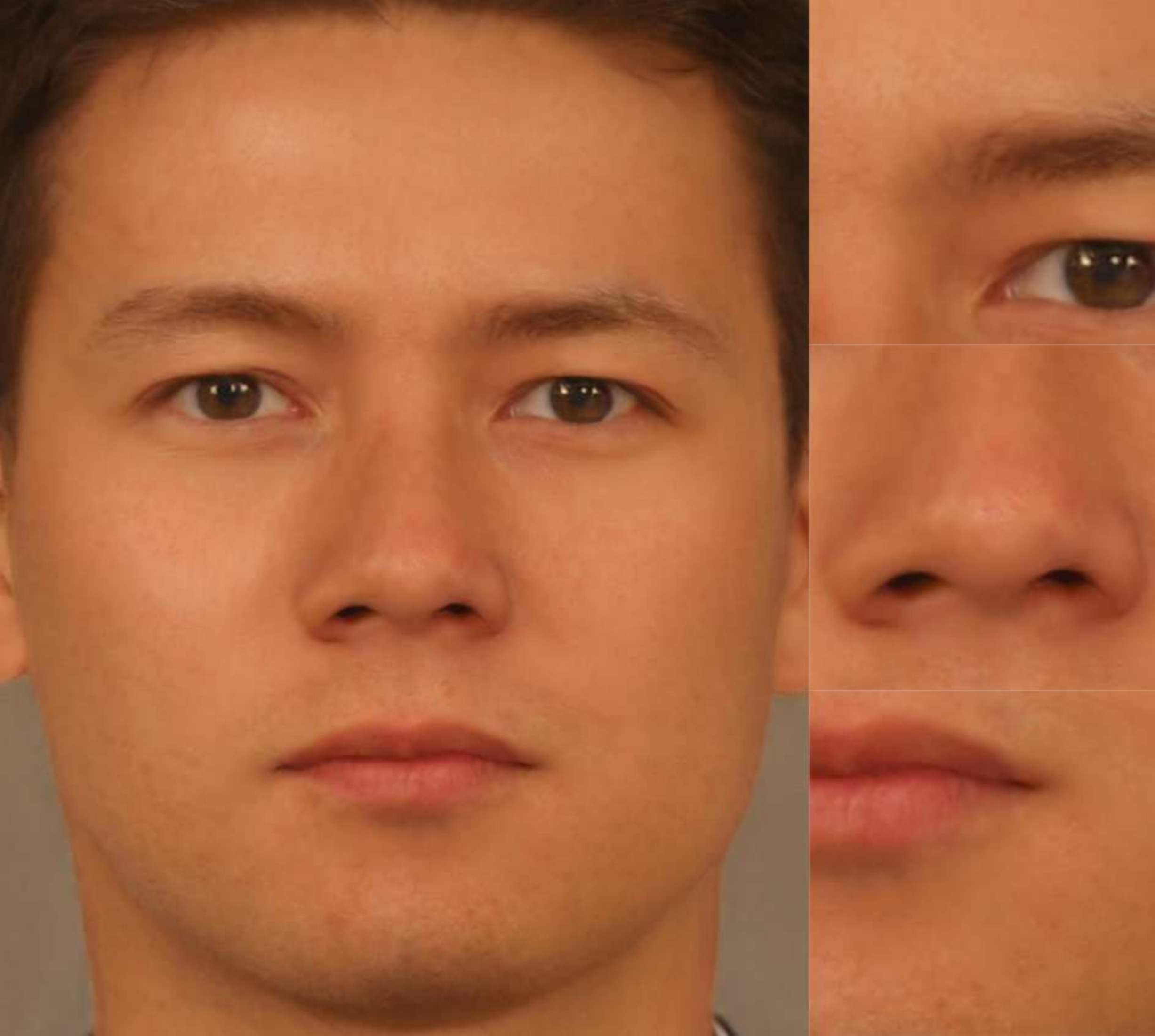}
          \\
         \includegraphics[height=\textwidth]{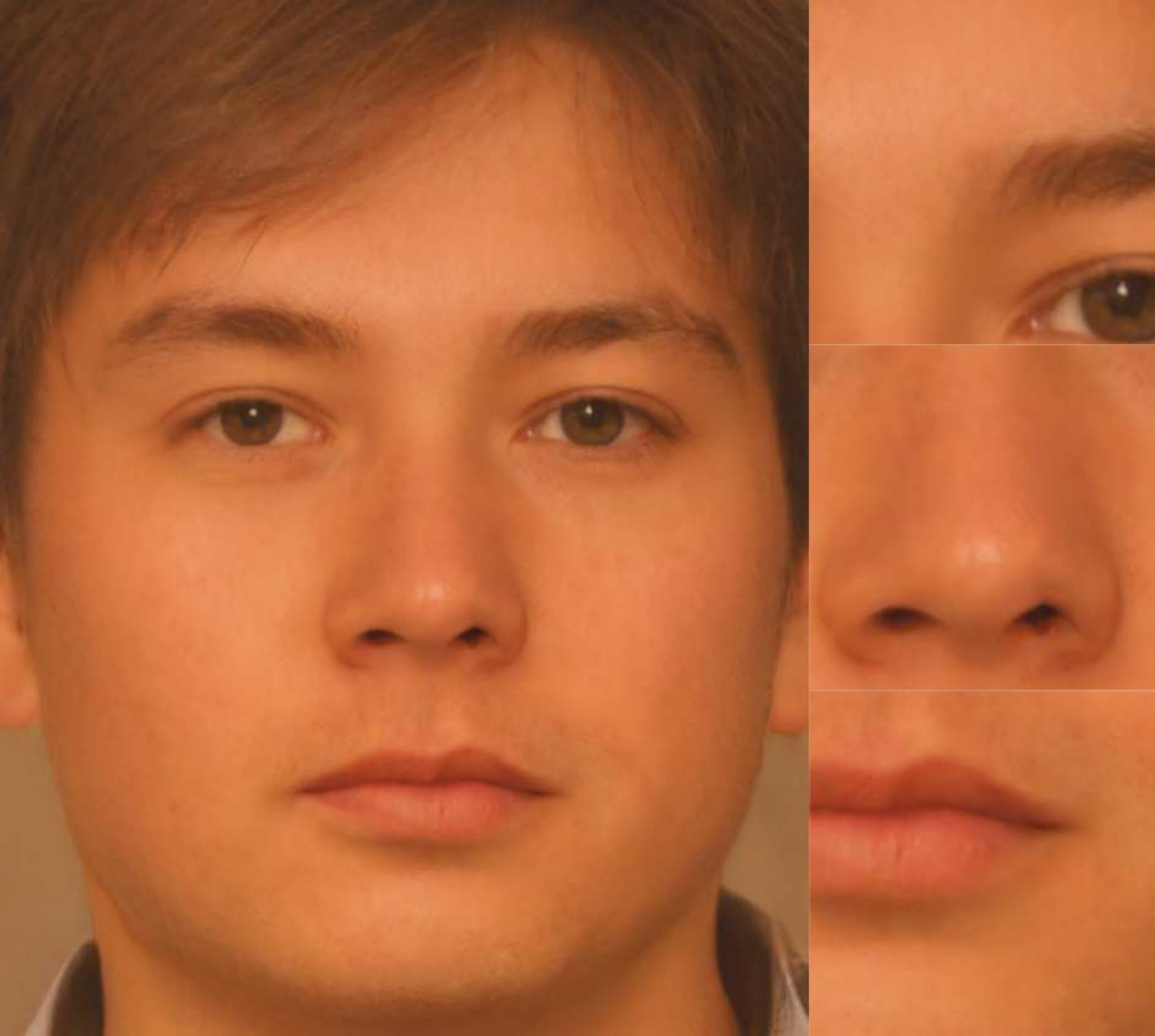}
         \\
         \includegraphics[height=\textwidth]{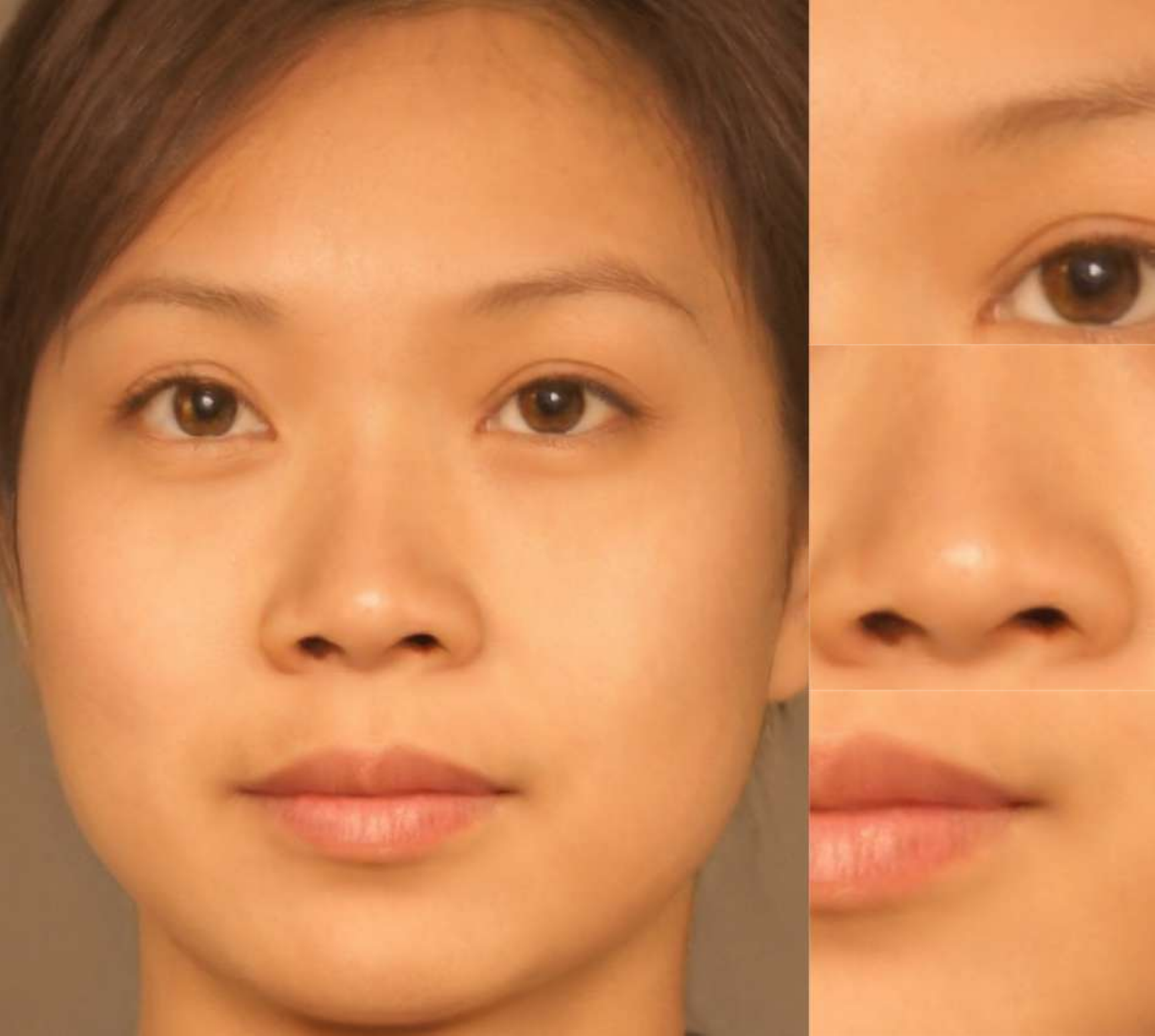}
         \\
         \includegraphics[height=\textwidth]{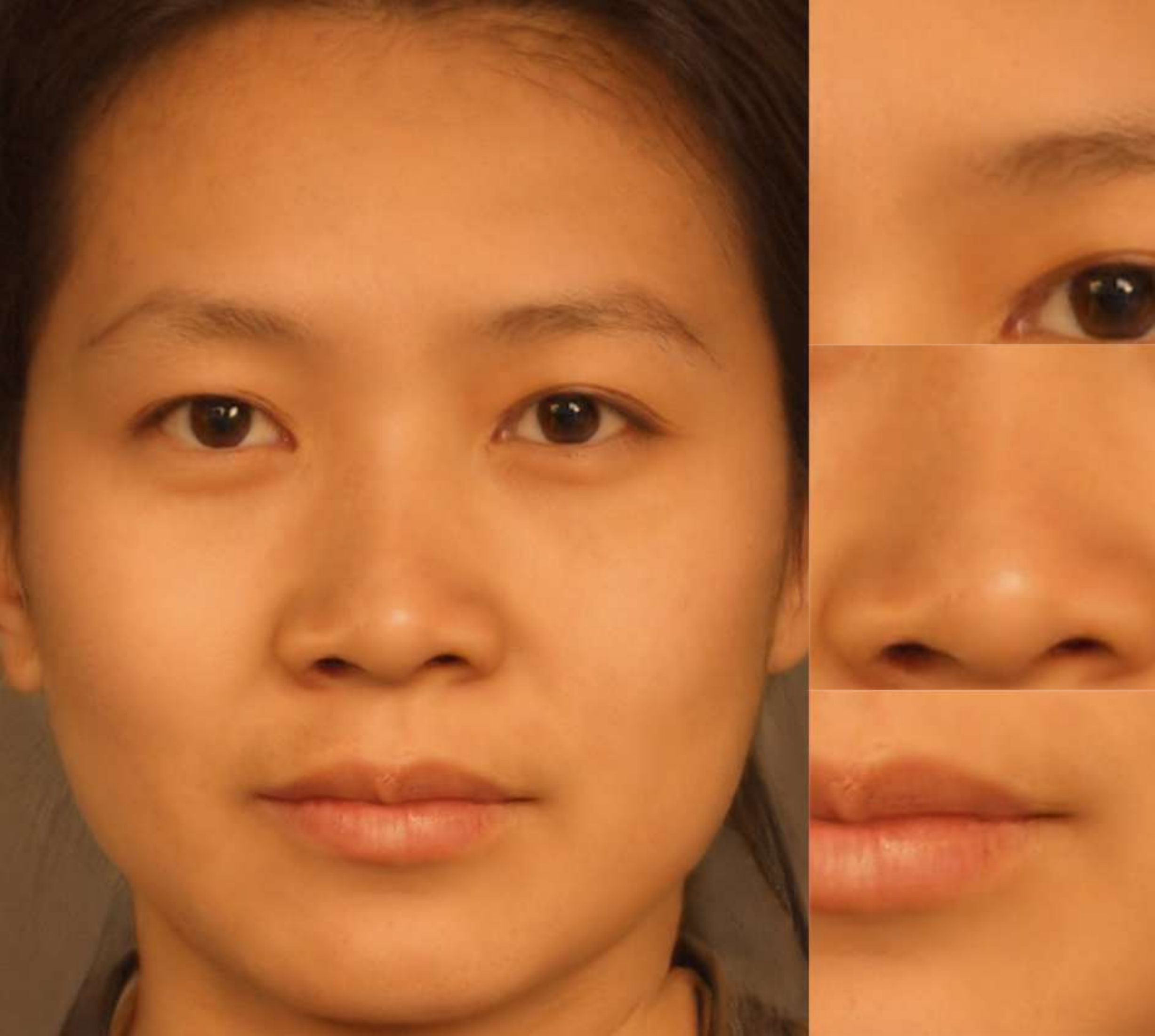}
         \caption{\tiny{ReGenMorph (ours)}}
         \label{fig:samp:BL}
     \end{subfigure}
        \caption{Samples of the morphed images using the LMA \cite{DBLP:conf/icb/RaghavendraRVB17}, StylGAN \cite{DBLP:conf/iwbf/VenkateshZRRDB20}, MIPGAN-II \cite{DBLP:journals/corr/abs-2009-01729MIPGAN}, and the proposed ReGenMorph approaches. The source image pairs that are used to create the morphs are on the far left. The zoomed areas on the right of each morphed image are indicated in the top LMA morphed image with red squares. One can notice the relatively more realistic nature of the ReGenMorphs. full scale examples are presented in Figure \ref{fig:samples_big}.}
        \label{fig:samples}
\end{figure*}

\section{Experimental Setup}
\label{sec:es}

This section presents the database used for the training and testing phase of the various investigation in this work.

\subsection{Database}
\label{sec:exp:db}

As there are no publically available morphing databases (to the best of our knowledge), and there are minimum private ones that adhere to ICAO standard \cite{ICAO}. To enhance comparison to state-of-the-art, we chose the largest ICAO compliant database in the literature, requested the database pairs details from the authors of the first published version \cite{DBLP:conf/iwbf/VenkateshZRRDB20}, and regenerated the images from the latest works \cite{DBLP:conf/icb/RaghavendraRVB17,DBLP:conf/iwbf/VenkateshZRRDB20,DBLP:journals/corr/abs-2009-01729MIPGAN}. We were not able to get access to the pairs and source image information of any other ICAO compliant \cite{ICAO} morphing data. However, we stress that the generalization is more important when presenting a morphing attack detector, while we present a novel attack generation protocol. Therefore one ICAO compliant database that enables a larger comparison with morphing approaches was our drive behind the database choice.
The used data is part of the FRGC-V2 face database\cite{DBLP:conf/cvpr/PhillipsFSBCHMMW05}. Based on this data, we generate morphing attacks using the proposed ReGenMorph approach, as well as the LMA approach defined in \cite{DBLP:conf/icb/RaghavendraRVB17}, the StyleGAN approach defined in \cite{DBLP:conf/iwbf/VenkateshZRRDB20}, and the MIPGAN-II approach defined in \cite{DBLP:journals/corr/abs-2009-01729MIPGAN}. 
For the sake of comparability, the generated database is based on the same database protocol and design (image pairs) as defined in, and provided by the authors of, \cite{DBLP:journals/corr/abs-2009-01729MIPGAN}. 
The database consists of 2500 morphed images of each of the four kinds (LMA, StyleGAN, MIPGAN-II, ReGenMorph) and 1270 bona fide images. These morphing attacks are split into identity-disjoint training and testing sets. The training set contains 1190 morphed images of each kind and 690 bona fide images. The testing set contains 1310 morphed images of each kind and 580 bona fide images. The training set is used to fine-tune the StyleGAN encoder as discussed in Section \ref{sec:meth:ft} and to train the MADs presented in Section \ref{sec:exp:det}. The testing set is used to show visual generation examples (Section \ref{res:vis}), study face recognition vulnerability to the considered attacks (Section \ref{res:vul}), and evaluate the attack detectability (Section \ref{res:det}).
Other GAN-based morphing techniques, such as the MorGAN \cite{DBLP:conf/btas/DamerS0K18} and EMorGAN \cite{DBLP:conf/btas/DamerBSKK19}, were not considered, as their quality and identity preservation ability is far inferior to recent GAN-based approaches (like StyleGAN \cite{DBLP:conf/iwbf/VenkateshZRRDB20} and MIPGAN \cite{DBLP:journals/corr/abs-2009-01729MIPGAN}). An LMA morphing tool from the University of Bologna \cite{DBLP:journals/tifs/FerraraFM18} is widely used, however, it is not considered here as it proved \cite{DBLP:journals/iet-bmt/ScherhagKRB20} to be inferior (in identity preservation) to the LMA approach used in this work \cite{DBLP:conf/icb/RaghavendraRVB17}.
We considered the option of measuring the statistical and perceptual image quality of the used morphs and found that previous works have shown no clear correlation between the image quality and the realistic appearance when dealing with Morphing attacks (typically of ICAO standard \cite{ICAO} with not large quality variation) \cite{DBLP:journals/corr/abs-2009-01729MIPGAN}. Therefore we decided that such investigation have proven to lead to misleading conclusions, and thus opted out of such investigation. 
The value of PSNR and SSIM, on the same morphing pairs used in this work, in \cite{DBLP:journals/corr/abs-2009-01729MIPGAN} showed insignificant difference between the different attacks and the little difference was inconsistent with the perceived visual quality. The work in \cite{DBLP:conf/btas/DamerBSKK19} showed that the clearly visibly unrealistic images of MorGAN \cite{DBLP:conf/btas/DamerS0K18} achieve higher statistical quality metrics (6 different metrics). In a recent study, researchers have shown that operations that apparent morphing artifacts do not consistently effect the estimated quality across a large number of quality estimation strategies \cite{DBLP:conf/biosig/FuMorph21}. Additionally, in a clearer study, Debiasi et al. have shown that even though MorGAN attacks have clearly low realistic appearance, they show closer perceptual quality distributions to bona fide images than attacks of the more realistic appearance.

\subsection{Vulnerability analyses}
\label{sec:exp:vul}

We evaluated the vulnerability of two face recognition systems to the morphing attacks created by our ReGenMorph pipeline in comparison to the three baseline attacks. In addition, we opted to evaluate one academic face recognition system and one commercial off-the-shelf (COTS) system. The following face recognition solutions were investigated:

\textbf{ArcFace:}  ArcFace \cite{DBLP:conf/cvpr/DengGXZ19} scored state-of-the-art performance on several face recognition evaluation benchmarks such as LFW $99.83\%$ and YTF $ 98.02\%$. ArcFace introduced Additive Angular Margin loss (ArcFace) to improve the discriminative ability of the face recognition model. We deployed ArcFace based on ReseNet-100 \cite{DBLP:conf/cvpr/HeZRS16} architecture pre-trained on a refined version of the MS-Celeb-1M dataset \cite{DBLP:conf/eccv/GuoZHHG16} (MS1MV2). The MTCNN \cite{zhang2016joint} solution is used, as recommended in \cite{DBLP:conf/cvpr/LiuWYLRS17}, to detect (crop) and align (affine transformation) the face before passing into the network. To perform a comparison, an Euclidean distance, as recommended in \cite{DBLP:conf/cvpr/DengGXZ19}, is calculated between two feature vectors produced from the network.

\textbf{COTS:} We used the Cognitec FaceVACS SDK version 9.4.2 \cite{cognitecfrssdk}. We chose this COTS as the SDK achieved one of the best performances in the recent NIST report addressing the performance of vendor face verification products \cite{nist2020}. The face quality threshold was set to zero for probes and references to minimize the impact of inherent quality metrics. The full process containing face detection, alignment, feature extraction, and matching is part of the COTS and thus we are not able to provide details of its algorithm. Comparing two faces by the COTS produces a similarity score.


\begin{figure*}[ht!]
	\centering
		\begin{subfigure}[h]{0.23\textwidth}
			\centering
			\resizebox{01\linewidth}{!}{
			\includegraphics{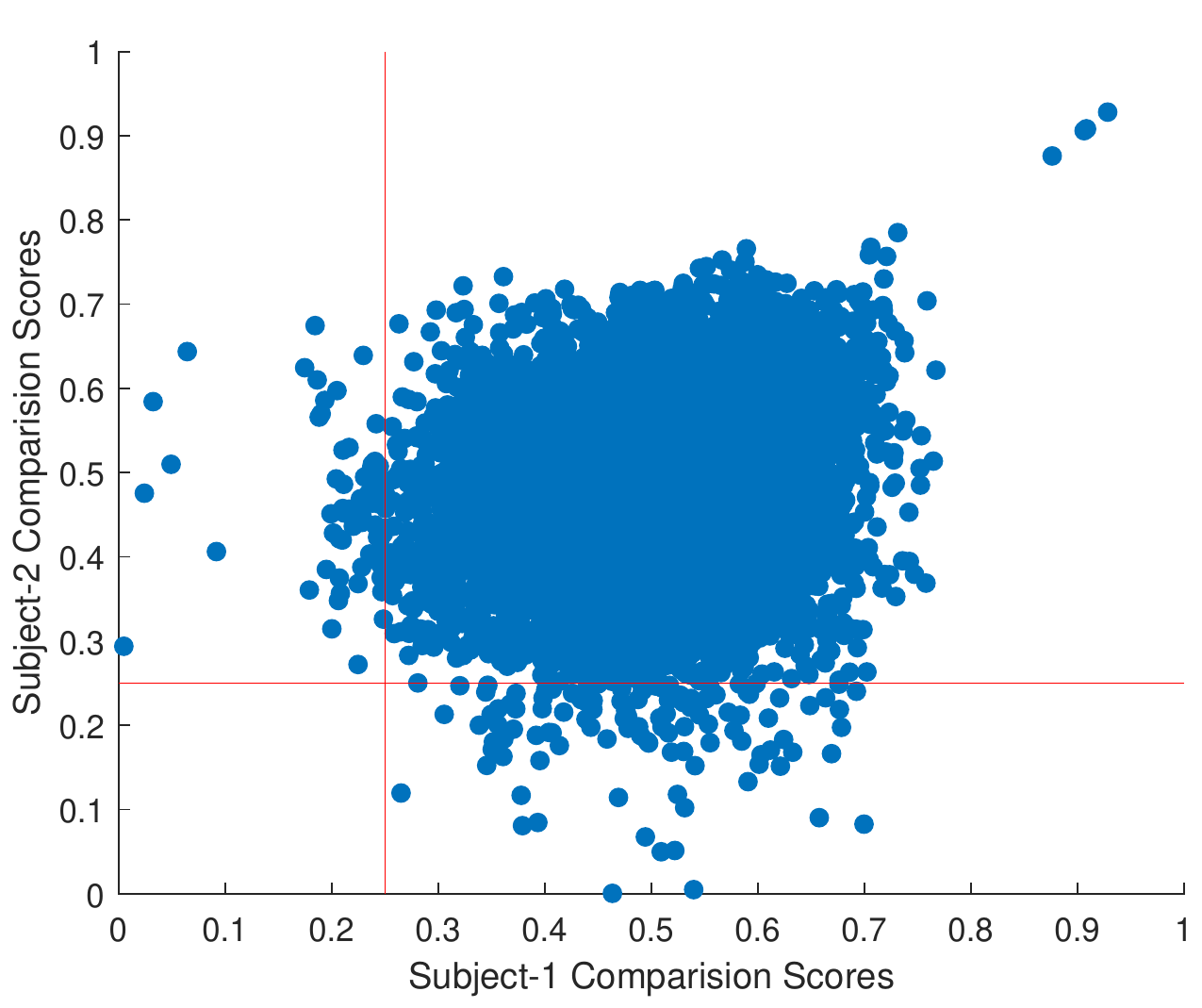}
			}
			\caption{LMA Arcface}
			\label{fig:LMA_ArcFace_updated}
		\end{subfigure}%
		\begin{subfigure}[h]{0.23\textwidth}
			\centering
			\resizebox{01\linewidth}{!}{
				\includegraphics{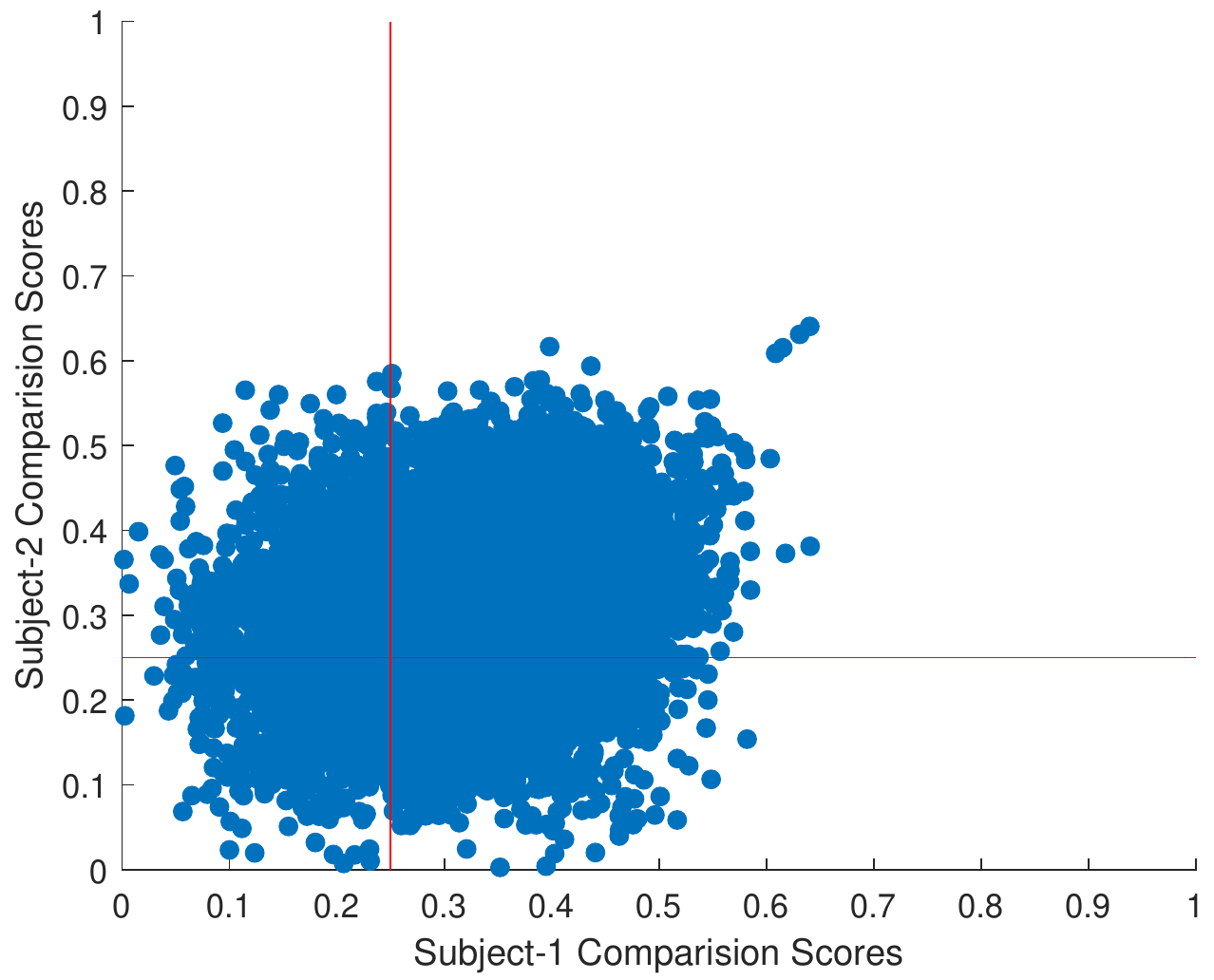}
			}
			\caption{StyleGAN - Arcface}
			\label{fig:StyleGAN_ArcFace_updated}
		\end{subfigure}%
		\begin{subfigure}[h]{0.23\textwidth}
			\centering
			\resizebox{01\linewidth}{!}{
				\includegraphics{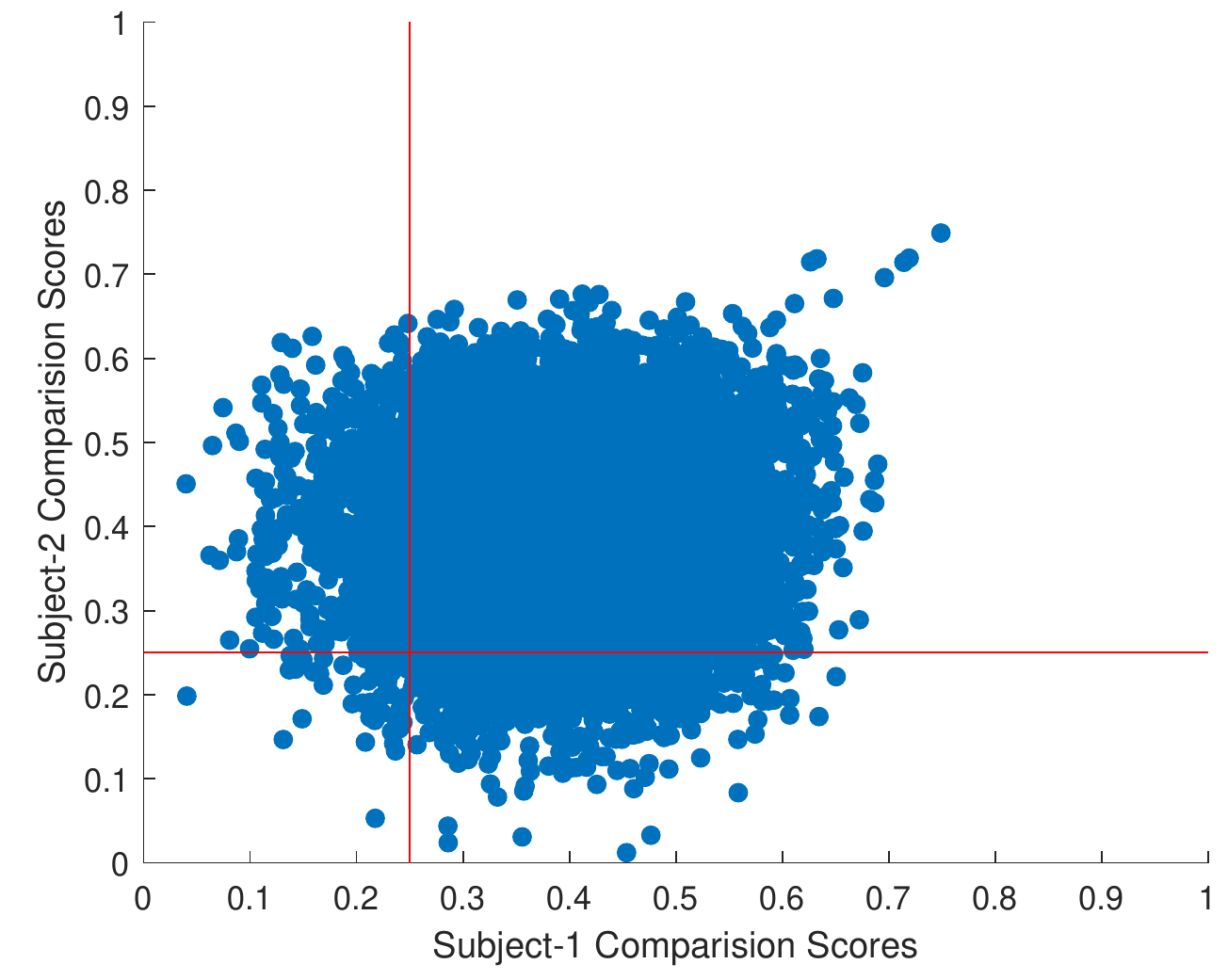}
			}
			\caption{MIPGAN-II - Arcface}
			\label{fig:StyleGAN_ArcFace_updated}
		\end{subfigure}%
		\begin{subfigure}[h]{0.23\textwidth}
			\centering
			\resizebox{01\linewidth}{!}{
				\includegraphics{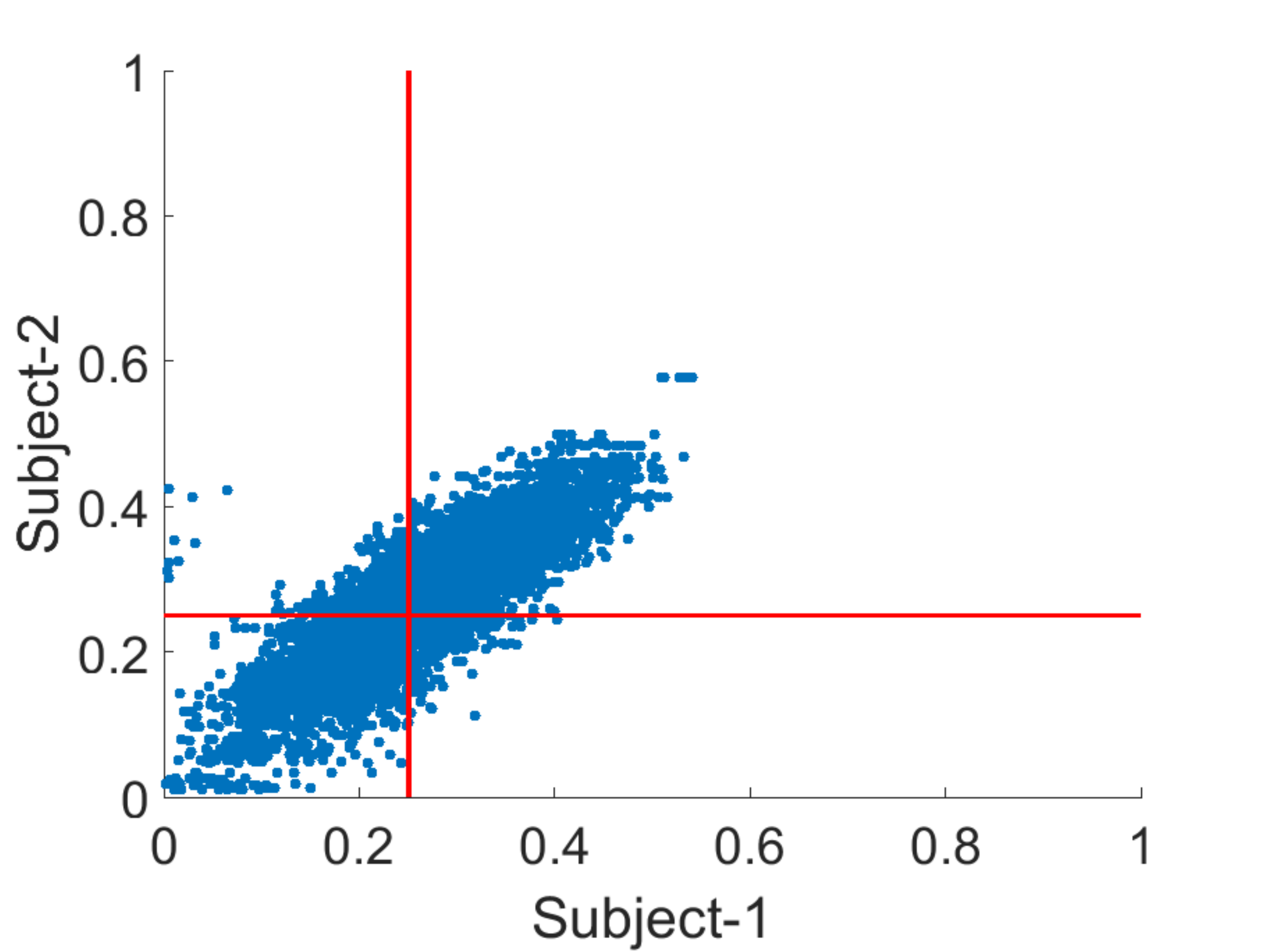}
			}
			\caption{ReGenMorph - Arcface}
			\label{fig:StyleGAN_ArcFace_updated}
		\end{subfigure}%
		\\
		\begin{subfigure}[h]{0.23\textwidth}
			\centering
			\resizebox{1\linewidth}{!}{
				\includegraphics{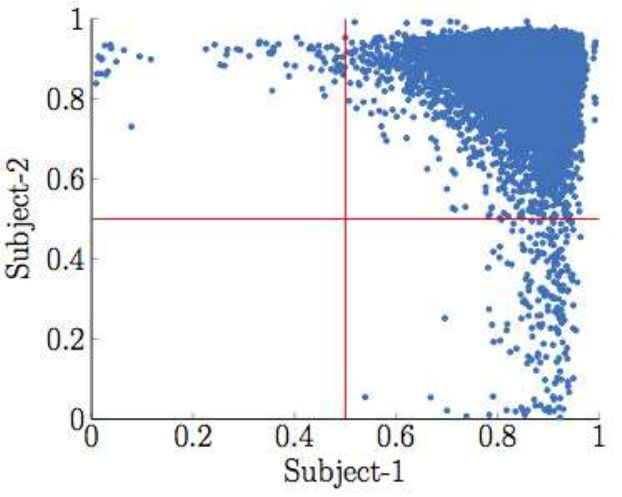}
			}
			\caption{LMA - COTS}
			\label{fig:LMA_cognitec}
		\end{subfigure}%
		\begin{subfigure}[h]{0.23\textwidth}
			\centering
			\resizebox{1\linewidth}{!}{
				\includegraphics{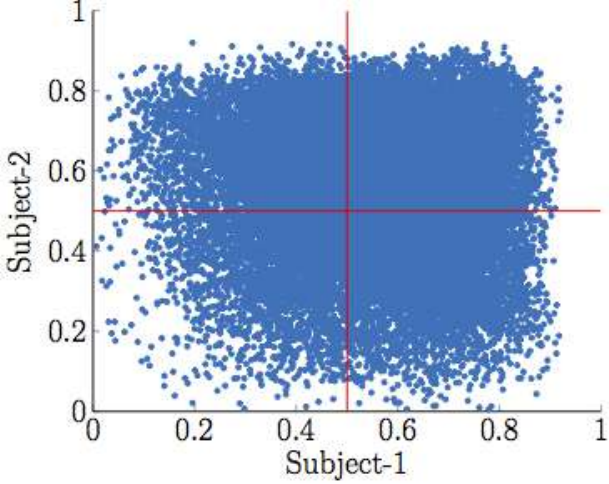}
			}
			\caption{StyleGAN - COTS}
			\label{fig:StyleGAN_Congitek}
		\end{subfigure}%
		\begin{subfigure}[h]{0.23\textwidth}
			\centering
			\resizebox{01\linewidth}{!}{
			\includegraphics{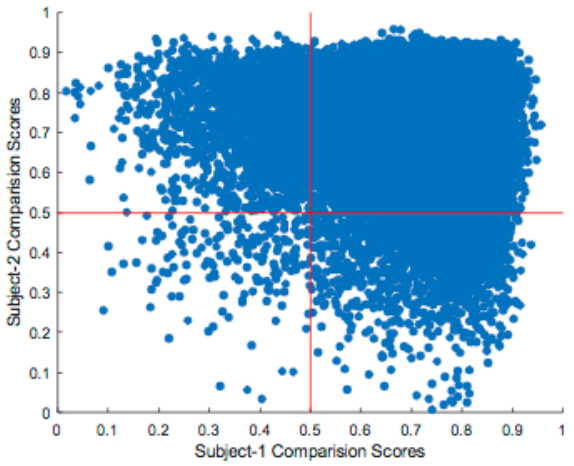}
			}
			\caption{MIPGAN-II - COTS}
			\label{fig:StyleGAN_ArcFace_updated}
		\end{subfigure}%
		\begin{subfigure}[h]{0.23\textwidth}
			\centering
			\resizebox{01\linewidth}{!}{
			\includegraphics{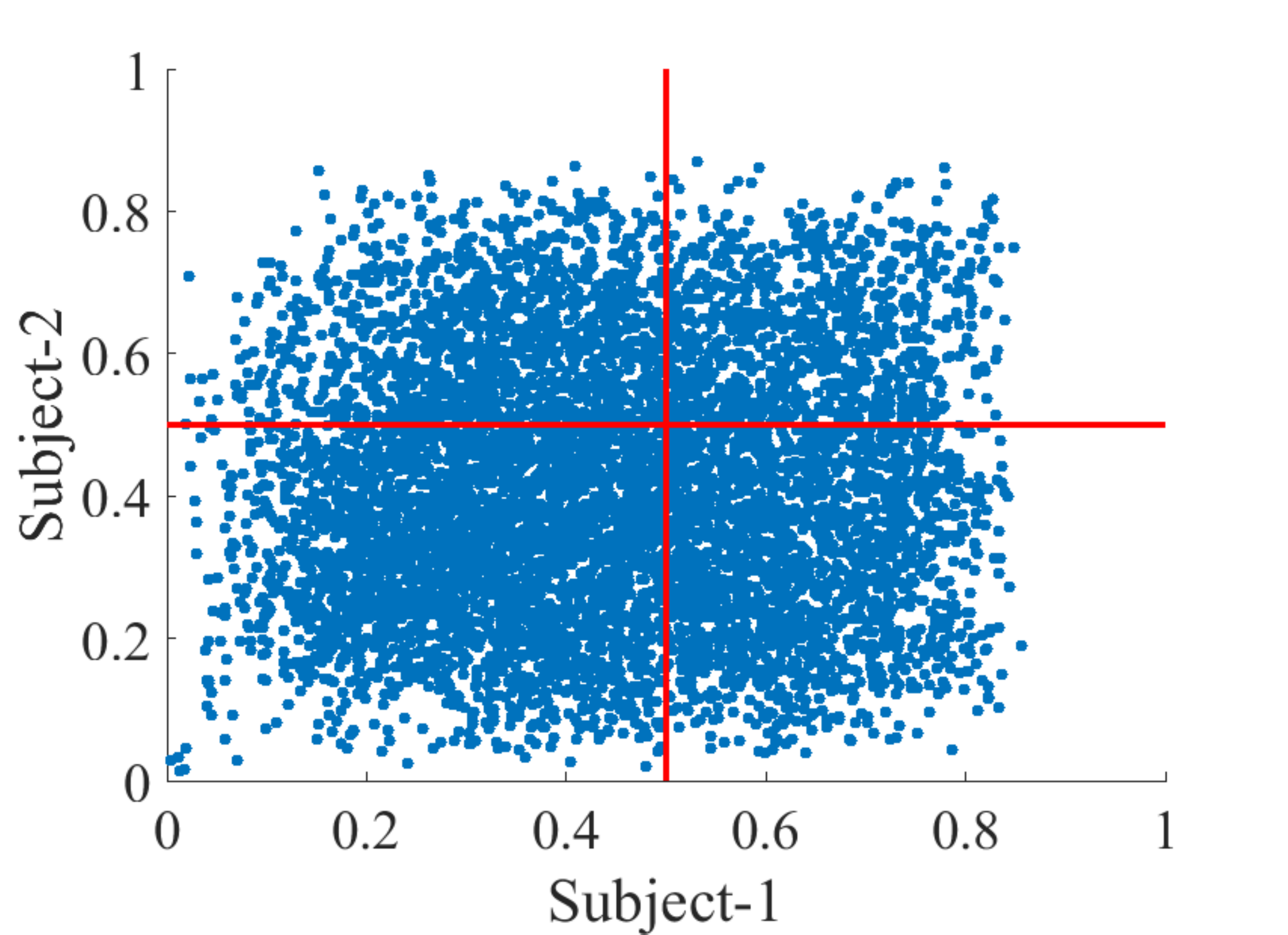}
			}
			\caption{ReGenMorph - COTS}
			\label{fig:StyleGAN_ArcFace_updated}
		\end{subfigure}%
	\caption{Vulnerability analysis using COTS and ArcFace. The scatter plots represents the comparison scores of morphed face image against two contributing subjects. The red lines represent decision threshold at 0.1\% FMR. The more diagonal and the more towards the top right corner the distribution is, the more vulnerable is the face recognition system to the attack.}
	\label{fig:vulnerability-from-iwbf}
\end{figure*}

We present the vulnerability study results in the visual form as scatter plots depicting the similarity score of each morphing attack image to reference images (not used to create the morphing image) of the target identities. Ideally, a strong morphing attack will have a similar, and high similarity score to the target identities. We also present the vulnerability results in a quantifiable manner by giving the Mated
Morphed Presentation Match Rate (MMPMR) \cite{DBLP:conf/biosig/ScherhagNRGVSSM17} and the Fully Mated Morphed Presentation Match Rate (FMMPMR) \cite{DBLP:conf/biosig/ScherhagNRGVSSM17} based on the decision threshold at the false match rate of (FMR) 0.1\%. It must be noted that all the vulnerability results are presented on the testing data.

\begin{table*}[]
\centering
\begin{tabular}{|c|c|c|c|c|}
\hline
\multirow{2}{*}{Generation}                                & \multicolumn{2}{c|}{COTS} & \multicolumn{2}{c|}{ArcFace} \\ \cline{2-5} 
                                                           & MMPMR(\%)       & FMMPMR(\%)      & MMPMR(\%)        & FMMPMR(\%)        \\ \hline
LMA \cite{DBLP:conf/icb/RaghavendraRVB17} & 100.00         & 98.84       & 99.68        & 98.00         \\ \hline
StyleGAN \cite{DBLP:conf/iwbf/VenkateshZRRDB20} & 64.68 & 41.49 & 72.80 & 56.95 \\ \hline
MIPGAN-II \cite{DBLP:journals/corr/abs-2009-01729MIPGAN}          & 92.93       & 81.59       & 94.21        & 86.94         \\ \hline
MorGAN \cite{DBLP:conf/btas/DamerS0K18,DBLP:conf/iwbf/VenkateshZRRDB20}                                                 & 0.00       & 0.00       & 0.00        & 0.00         \\ \hline
ReGenMorph                                                 & 42.24       & 34.47       & 33.98        & 14.05         \\ \hline
\end{tabular}%
\caption{Quantitative evaluation of vulnerability of COTS and ArcFace to various morph generation approaches. All values are in percentage (\%). All value from our presented data, except for MorGAN, reported from the same database protocol in \cite{DBLP:conf/iwbf/VenkateshZRRDB20}.}
\label{tab:vulnerability-cots-arcface}
\end{table*}

\subsection{Detectability analyses}
\label{sec:exp:det}

To measure the detectability of the ReGenMorph attacks, we deploy two best-performing MADs. These two approaches are chosen by considering two different properties. One is based on Hybrid features \cite{DBLP:conf/isba/RamachandraVRB19} and the other based on Ensemble features \cite{DBLP:conf/fusion/VenkateshRRB20}.

The Hybrid features \cite{DBLP:conf/isba/RamachandraVRB19} extracts features using both scale space and color space combined with multiple classifiers, while the Ensemble features \cite{DBLP:conf/fusion/VenkateshRRB20} deploys different textural features in conjunction with a set of classifiers. Both approaches proved to detect variations of morphing attacks holistically as previously shown in \cite{DBLP:conf/isba/RamachandraVRB19,DBLP:conf/fusion/VenkateshRRB20}. In addition, the Hybrid features \cite{DBLP:conf/isba/RamachandraVRB19} is submitted to the ongoing NIST FRVT morph challenge dataset \cite{nistMorph} where it scored the best performance in detecting printed and scanned morph attacks.

We evaluate the detectability of our ReGenMorph attacks when they are already known to the detector developer, i.e. the training attack data is also ReGenAttack. More importantly, we evaluate the detectability of our ReGenMorph attacks as unknown attacks, i.e. novel attacks unknown to the detection algorithm. In the latter case, we evaluate the cases where the detectors are trained using LMA, StyleGAN, or MIPGAN-II attacks.  

The morphing attack performance (detectability) is presented by the Attack Presentation Classification Error Rate (APCER), i.e. the proportion of attack images incorrectly classified as bona fide samples, and the Bona fide Presentation Classification Error Rate (BPCER), i.e. the proportion of bona fide images incorrectly classified as attack samples, as defined in the ISO/IEC 30107-3 \cite{ISO301073}. Additionally, the Detection Equal Error Rate (D-EER), i.e. the value of APCER or BPCER at the decision threshold where they are equal, is reported. It must be noted that the training used only the training data while the detectability evaluation was performed only on the disjoint test data as described in \ref{sec:exp:db}.



\begin{table}[]
\centering
\begin{tabular}{|l|l|l|l|}
\hline
Approach                                                                  & EER (\%) & \multicolumn{2}{l|}{BPCER (\%) @ APCER} \\ \cline{3-4} 
                                                                          &          & =5\%               & =10\%              \\ \hline
Hybrid features\cite{DBLP:conf/isba/RamachandraVRB19}    & 2.48     & 4.97               & 4.97               \\ \hline
Ensemble features \cite{DBLP:conf/fusion/VenkateshRRB20} & 0.00     & 0.00               & 0.00               \\ \hline
\multicolumn{4}{|c|}{Cross-set - Trained on LMA}                                                                               \\ \hline
Hybrid features\cite{DBLP:conf/isba/RamachandraVRB19}    & 0.08     & 0.17               & 0.27               \\ \hline
Ensemble features \cite{DBLP:conf/fusion/VenkateshRRB20} & 0.16     & 0.17               & 0.17               \\ \hline
\multicolumn{4}{|c|}{Cross-set - Trained on MIPGAN-II}                                                                            \\ \hline
Hybrid features\cite{DBLP:conf/isba/RamachandraVRB19}    & 50.00    & 100.00             & 100.00             \\ \hline
Ensemble features \cite{DBLP:conf/fusion/VenkateshRRB20} & 33.34    & 70.33              & 82.68              \\ \hline
\end{tabular}
\caption{The Detectability of the ReGenMorph attacks when they are used for training the MAD (known) and when LMA or MIPGAN-II attacks are used for training the MAD. One can notice the more challenging unknown ReGenMorph attack scenario. All BPCER and EER values are in percentage (\%).}
\label{tab:mad-evaluation}
\end{table}



\section{Results}
\label{sec:res}

This section presents a visible representation of the ReGenMorph, a face recognition vulnerability analysis of the presented attacks, and detectability analyses of the ReGenMorph. 

\subsection{ReGenMorph image appearance}
\label{res:vis}

The visual appearance of morphed images is important because in the cases of 1) manual identity verification (e.g. border check) and 2) applying for an identity document with printed photo and no automatic attack detection, the only morphing detection possible is by the human operator. This human operator can commonly only identify visible image artifacts. 
Figure \ref{fig:samples} presents several morphed images created by the proposed ReGenMorph and the baselines LMA \cite{DBLP:conf/icb/RaghavendraRVB17}, StyleGAN \ref{fig:vulnerability-from-iwbf}, and MIPGAN-II \cite{DBLP:journals/corr/abs-2009-01729MIPGAN}. On the far left, the source images used to create the morphs are illustrated. It can be seen that the visible identity similarity of all morphed images to the source identities. LMA morphs show blending artifacts, especially in the nose and eye regions. The StyleGAN and MIPGAN-II morphs look slightly more synthetic and contain striping artifacts, especially in the flat skin areas. These artifacts are more visible in the MIPGAN-II, in comparison to StyleGAN morphs. There are significantly fewer artifacts in the proposed ReGenMorph morphs, with no clear consistent pattern of artifacts.
To take a clearer look, three areas of each morphed image are zoomed in, as clarified in the top LMA morphed image in Figure \ref{fig:samples}. To take a holistic clear look at the generated images, a morphing example from all the 4 morphing methods is presented in Figure \ref{fig:samples_big}. The figure shows the realistic and low-artifact appearance of the ReGenMorph compared with baselines.

\subsection{Vulnerability of face recognition to ReGenMorph}
\label{res:vul}

As detailed in Section \ref{sec:exp:vul}, we measure the vulnerability of one COTS and the ArcFace \cite{DBLP:conf/cvpr/DengGXZ19}. The MMPMR and FMMPMR at 0.1\% FMR are presented in Table \ref{tab:vulnerability-cots-arcface}. Both face recognition systems are less vulnerable to the ReGenMorph than to LMA, StyleGAN, or MIPGAN-II. However, many of the samples can cross the 0.1\% FMR. This makes the ReGenMorph attacks of interest, given the worst-case attack scenario, and the realistic visible appearance of the attacks. A similar conclusion can be made from the scatter plots in Figure \ref{fig:vulnerability-from-iwbf}. Where despite samples of ReGenMorph falling below the 0.1\% FMR similarity threshold, those who don't can be considered a serious threat, especially given the realistic image appearance. 
In a conclusion, the ReGenMorphs can be considered a serious threat to face recognition systems with MMPMR and FMMPMR over 30\% on top-performing COTS and at 0.1\%FMR.
Generally, as an attack, the system operators should not only be considering the strongest attack but should foresee all plausible attacks that can generate a serious threat. This is said, keeping in mind the importance of the realistic appearance as stated in Section \ref{res:vis}.

\subsection{Detectability of ReGenMorph}
\label{res:det}
As detailed in Section \ref{sec:exp:det}, We measure the detectability of the ReGenMorph attacks under the know and unknown attack scenarios by using two state-of-the-art MAD methods \cite{DBLP:conf/isba/RamachandraVRB19,DBLP:conf/fusion/VenkateshRRB20}. Table \ref{tab:mad-evaluation} presents the BPCER values at different APCER thresholds. It should be noticed that the ReGenMorph attacks can be efficiently detected when the MAD is trained on the ReGenMorph or the LMA attacks. The fact that the ReGenMorph attacks can be detected by MAD trained on LMA attack does not mean that both attacks have the same appearance artifacts, as this is shown not to be the case (see Figure \ref{fig:samples}). 
Detection algorithm can build their decisions on deeper statistical co-relations in the image that may not reflect the appearance, an example of that are the numerous studies on adversarial attacks \cite{DBLP:journals/cviu/MassoliCAF21}.
On the other hand, the MAD performance is significantly worse when trained on the MIPGAN-II attacks. Both investigated MAD methods tend to have the same behavior pattern, however with varying performances (see Table \ref{tab:mad-evaluation}). Beyond the automatic detectability, we stress that the realistic visual appearance is a serious threat to manual identity inspection as mentioned in Section \ref{res:vis}. Therefore, it is essential to foresee new morphing concepts as the one presented in this work.

\section{Conclusion}
\label{sec:con}

We propose in this work the novel ReGenMorph morphing pipeline to avoid blending artifacts caused by image-level morphing in LMA and synthetic-like striping artifacts caused by latent vector manipulation in GAN-based morphs. ReGenMorph exploits the GAN capabilities of creating realistic images without manipulating the latent vector, by feeding the encoder with an image that is morphed on the image level. The generated morph images exhibit high appearance quality in comparison to the latest morphing approaches. The identity preservation of the ReGenMorph is lower than the top morphing methods, however, given the security nature of the attack, worse-case scenarios should be considered. The detectability of the ReGenMorphs is low when they are not considered in the training, especially when the MAD is trained on the recent MIPGAN-II attacks.

\paragraph{Acknowledgment:} This research work has been funded by the German Federal Ministry of Education and Research and the Hessian Ministry of Higher Education, Research, Science and the Arts within their joint support of the National Research Center for Applied Cybersecurity ATHENE.

%
%
%



\section*{Supplementary material}

# Supplementary Material
## ReGenMorph: High Quality GAN Generated Face Morphing Attacks by Attack Re-generation

Naser Damer[1,2], Kiran Raja[3], Marius Süßmilch[1], Sushma Venkatesh[3], Fadi Boutros[1,2], Meiling Fang[1,2], Florian Kirchbuchner[1], Raghavendra Ramachandra[3], and Arjan Kuijper[1,2]

[1] Fraunhofer Institute for Computer Graphics Research IGD, Germany
[2] Department of Computer Science, TU Darmstadt, Germany
[3] Norwegian University of Science and Technology, Norway
naser.damer@igd.fraunhofer.de

This supplementary material is presented to enable having an easier perception of the visual quality of the presented ReGenMorph attacks in comparison to the top quality GAN-based morphing attacks (i.e. StyleGAN [2] and MIPGAN-II [3]) and the more conventional landmark-based (LMA) attacks [1]. The presented images are provided in high resolution in the paper, however, they require the reader to zoom into the images to get a better visual evaluation, therefore, to enable an easier review process, we provide the images in their full scale in Figures 1, 2, 3, 4, 5, 6, 7, and 8.

Please note in all the figures the following:

– The LMA attacks typically contain blending artifacts appearing especially in detailed areas, such as the nose and the eyebrows.
– The StyleGAN and MIPGAN-II attacks contain striping artifacts, especially in the continuous skin areas. This renders the image to appear slightly artificial. these effects are generally more visible in the MIPGAN-II images in comparison to the StyleGAN ones.
– Both types or artifacts appear to be minimized in our ReGenMorph attacks. The viewer can notice the less artificial appearance of the ReGenMorph images in comparison to StyleGAN and MIPGAN-II images.

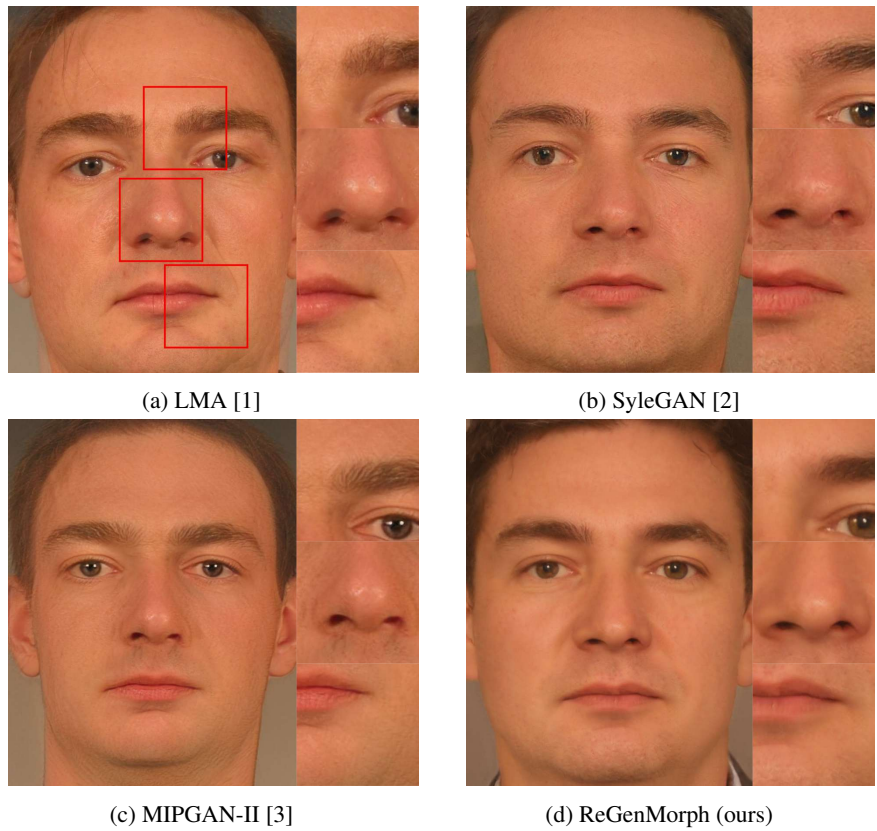

(a) LMA [1]  (b) SyleGAN [2]

(c) MIPGAN-II [3]  (d) ReGenMorph (ours)

Fig. 1: Samples of a morphed image using LMA [1], StyleGAN [2], MIPGAN-II [3], and our ReGenMorph.

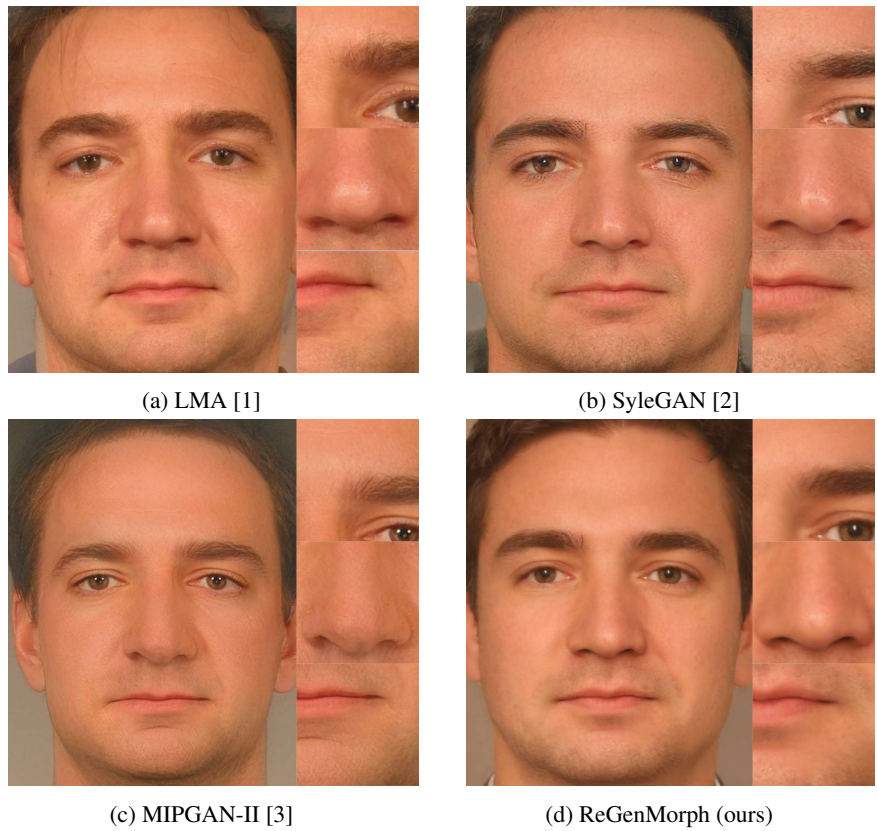

Fig. 2: Samples of a morphed image using LMA [1], StyleGAN [2], MIPGAN-II [3], and our ReGenMorph.

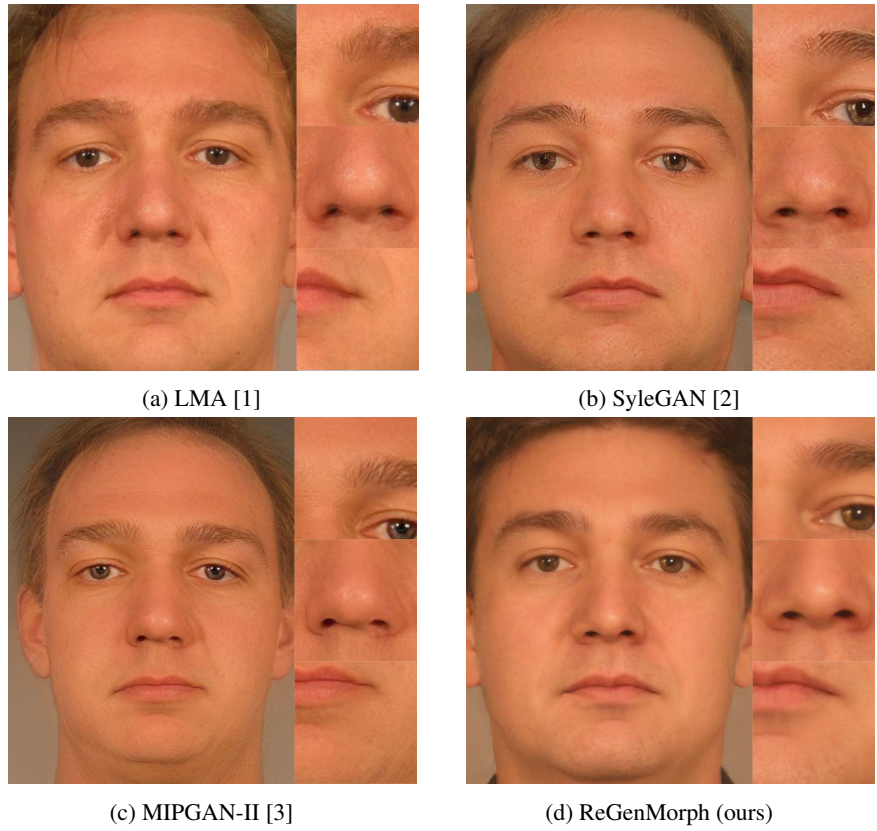

(a) LMA [1]  (b) SyleGAN [2]

(c) MIPGAN-II [3]  (d) ReGenMorph (ours)

Fig. 3: Samples of a morphed image using LMA [1], StyleGAN [2], MIPGAN-II [3], and our ReGenMorph.

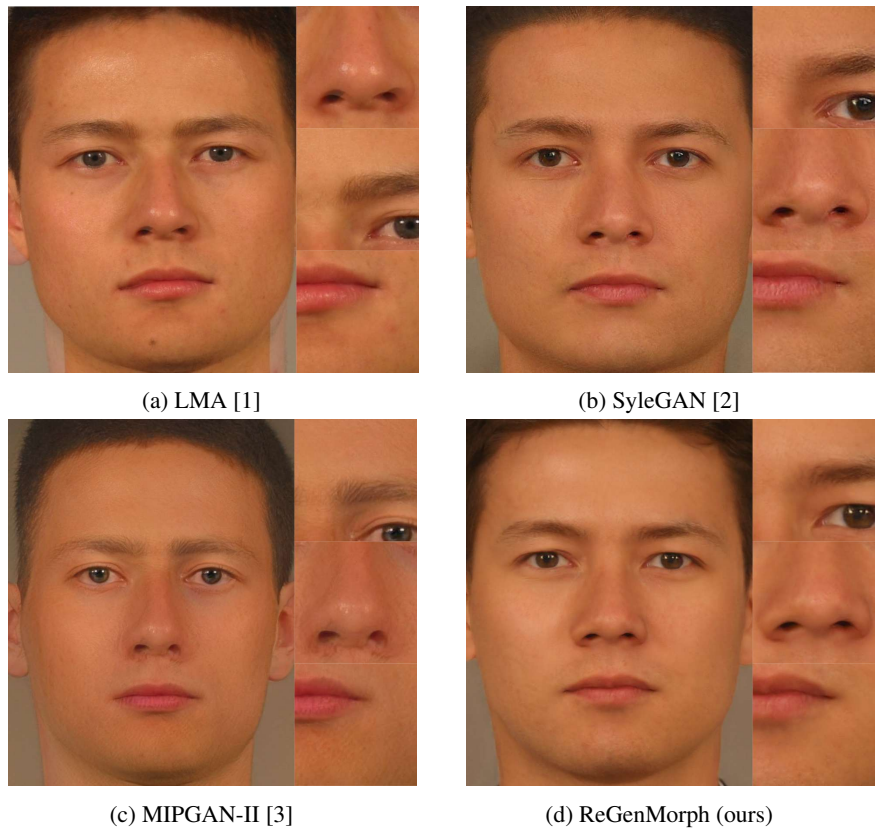

(a) LMA [1]        (b) SyleGAN [2]

(c) MIPGAN-II [3]        (d) ReGenMorph (ours)

Fig. 4: Samples of a morphed image using LMA [1], StyleGAN [2], MIPGAN-II [3], and our ReGenMorph.

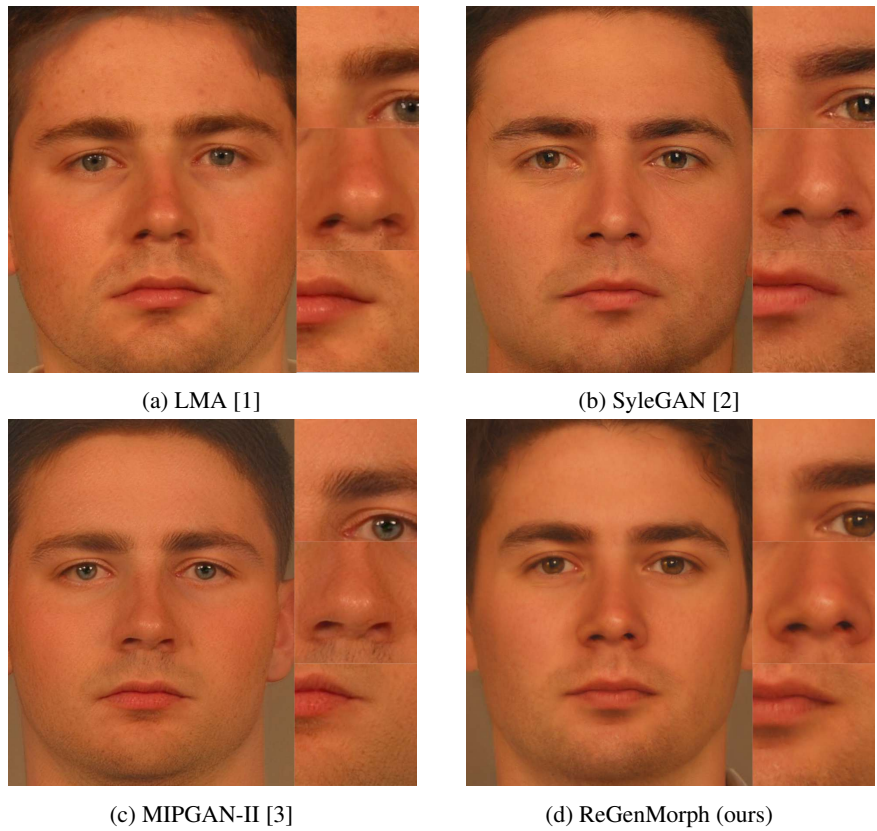

(a) LMA [1]   (b) SyleGAN [2]

(c) MIPGAN-II [3]   (d) ReGenMorph (ours)

Fig. 5: Samples of a morphed image using LMA [1], StyleGAN [2], MIPGAN-II [3], and our ReGenMorph.

ReGenMorph - Supplementary Material 7

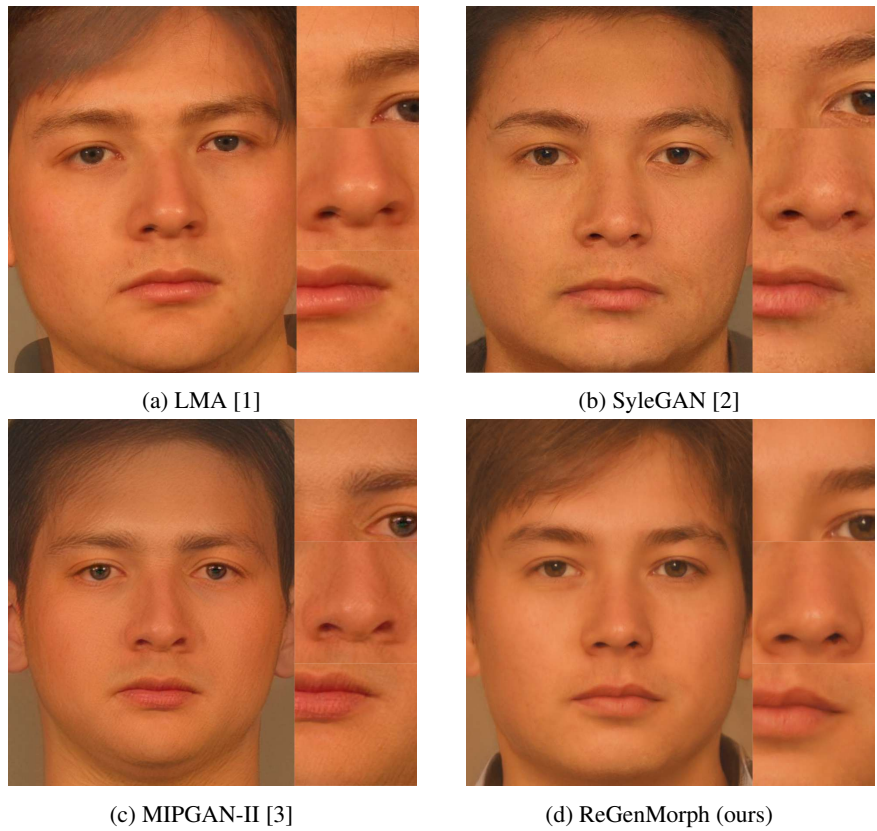

Fig. 6: Samples of a morphed image using LMA [1], StyleGAN [2], MIPGAN-II [3], and our ReGenMorph.

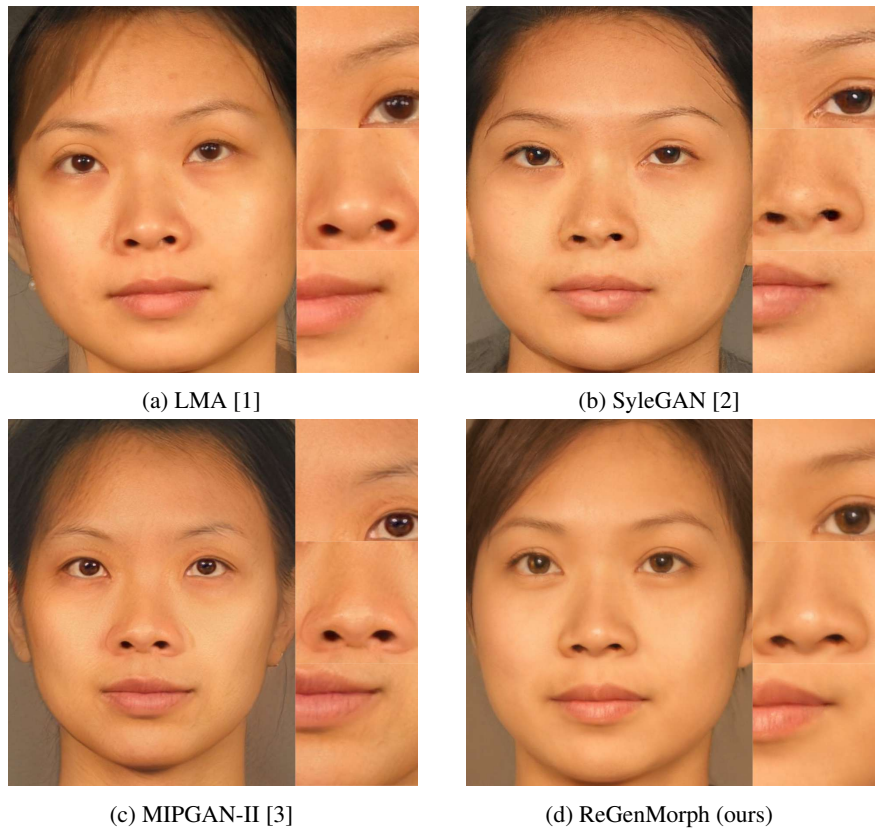

(a) LMA [1]  (b) SyleGAN [2]

(c) MIPGAN-II [3]  (d) ReGenMorph (ours)

Fig. 7: Samples of a morphed image using LMA [1], StyleGAN [2], MIPGAN-II [3], and our ReGenMorph.

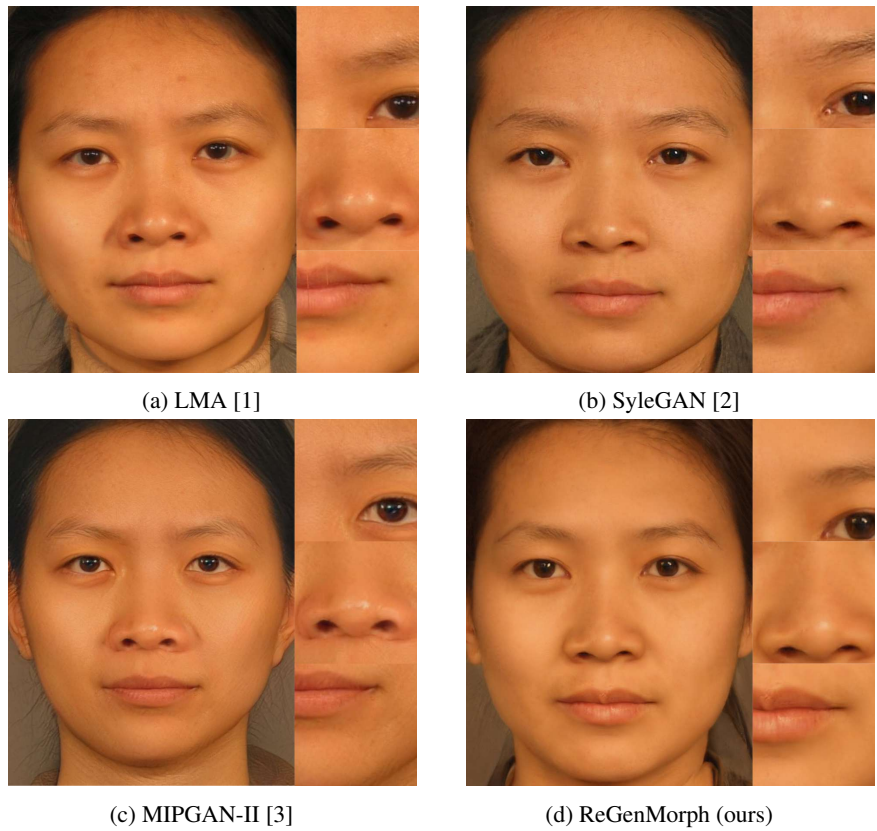

(a) LMA [1]  (b) SyleGAN [2]

(c) MIPGAN-II [3]  (d) ReGenMorph (ours)

Fig. 8: Samples of a morphed image using LMA [1], StyleGAN [2], MIPGAN-II [3], and our ReGenMorph.



\bibliographystyle{splncs04}
\bibliography{main}

\end{document}